\documentclass{ieeeaccess}
\usepackage{cite}
\usepackage{amsmath,amssymb,amsfonts}
\usepackage{dsfont}
\usepackage{algorithmic}
\usepackage{graphicx}
\usepackage{subfigure}
\usepackage{textcomp}
\usepackage{multirow}
\usepackage{tabularx}
\usepackage{wrapfig}
\usepackage{rotating}
\usepackage[export]{adjustbox}
\usepackage[strict]{changepage}
\usepackage{float} 
\usepackage{placeins}
\usepackage{hyperref}

\usepackage{caption,setspace} 

\newcommand{\x}{{\bf x}}
\newcommand{\y}{{\bf y}}

\newcommand{\beq}{\begin{equation}}
\newcommand{\eeq}{\end{equation}}
\newcommand{\bde}{\begin{definition}}
\newcommand{\ede}{\end{definition}}
\newcommand{\bpp}{\begin{property}}
\newcommand{\epp}{\end{property}}
\newcommand{\bpr}{\begin{proposition}}
\newcommand{\epr}{\end{proposition}}
\newcommand{\bex}{\begin{example}}
\newcommand{\eex}{\end{example}}
\newcommand{\bco}{\begin{corollary}}
\newcommand{\eco}{\end{corollary}}
\newcommand{\bre}{\begin{remark}}
\newcommand{\ere}{\end{remark}}
\newcommand{\bal}{\begin{algorithm}}
\newcommand{\eal}{\end{algorithm}}
\newcommand{\ble}{\begin{lemma}}
\newcommand{\ele}{\end{lemma}}

\definecolor{amber(sae/ece)}{rgb}{1.0, 0.49, 0.0}
\definecolor{amberlite(sae/ece)}{rgb}{1.0, 0.69, 0.2}
\newcommand{\revised}[1]{{\color{amber(sae/ece)}{#1}}}
\newcommand\Tstrut{\rule{0pt}{2.8ex}} 

\def\BibTeX{{\rm B\kern-.05em{\sc i\kern-.025em b}\kern-.08em
    T\kern-.1667em\lower.7ex\hbox{E}\kern-.125emX}}
\begin{document}
\history{Date of publication xxxx 00, 0000, date of current version xxxx 00, 0000.}
\doi{xx.xxxx/ACCESS.2021.DOI}

\title{Comparison of single and multitask 
learning for predicting cognitive decline based on MRI data}
\author{\uppercase{Vandad Imani}\authorrefmark{1}\href{https://orcid.org/0000-0003-0198-3400}{\includegraphics[scale=0.07]{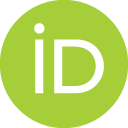}}, 
\uppercase{Mithilesh Prakash}\authorrefmark{1}\href{https://orcid.org/0000-0002-3853-4126}{\includegraphics[scale=0.07]{ORCIDiD_icon128x128.png}},\uppercase{Marzieh Zare} \authorrefmark{2}\href{https://orcid.org/0000-0003-0229-9002}{\includegraphics[scale=0.07]{ORCIDiD_icon128x128.png}} \uppercase{and, Jussi Tohka}\authorrefmark{1}\href{https://orcid.org/0000-0002-1048-5860}{\includegraphics[scale=0.07]{ORCIDiD_icon128x128.png}}, Alzheimer’s Disease Neuroimaging Initiative.}

\address[1]{A. I. Virtanen Institute for Molecular Sciences, University of Eastern Finland, 70211 Kuopio, Finland (e-mail: author@uef.fi)}
\address[2]{Department of Computing Sciences, Tampere University, Finland. (e-mail: author@tuni.fi)}

\tfootnote{This work was supported in part by the Academy of Finland under Grant 316258. Data used in preparation of this article were obtained from the Alzheimer’s Disease
Neuroimaging Initiative (ADNI) database (adni.loni.usc.edu). As such, the investigators within the ADNI contributed to the design and implementation of the ADNI and/or provided data but did not participate in the analysis or writing of this report. A complete listing of ADNI investigators can be found at:
http://adni.loni.usc.edu/wp-content/uploads/how\_to\_apply/ADNI\_Acknowledgement\_List.pdf}

\markboth
{V. Imani \headeretal: Comparison of single and multitask learning for predicting cognitive decline based on MRI data}
{V. Imani \headeretal: Comparison of single and multitask learning for predicting cognitive decline based on MRI data}

\corresp{Corresponding author: Vandad Imani (e-mail: vandad.imani@uef.fi).}

\begin{abstract}
The Alzheimer's Disease Assessment Scale-Cognitive subscale (ADAS-Cog) is a neuropsychological tool that has been designed to assess the severity of cognitive symptoms of dementia. Personalized prediction of the changes in ADAS-Cog scores could help in timing therapeutic interventions in dementia and at-risk populations. In the present work, we compared single and multitask learning approaches to predict the changes in ADAS-Cog scores based on T1-weighted anatomical magnetic resonance imaging (MRI). In contrast to most machine learning-based prediction methods 
ADAS-Cog changes, we stratified the subjects based on their baseline diagnoses and evaluated the prediction performances in each group. Our experiments indicated a positive relationship between the predicted and observed ADAS-Cog score changes in each diagnostic group, suggesting that T1-weighted MRI has a predictive value for evaluating cognitive decline in the 
entire AD continuum. We further studied whether correction of the differences in the magnetic field strength of MRI would improve the ADAS-Cog score prediction. The partial least square-based domain adaptation 
slightly improved the prediction performance, but the improvement was marginal. In summary, this study demonstrated that ADAS-Cog change 
could be, to some extent, predicted based on anatomical MRI. Based on this study, the recommended method for learning the predictive models is a single-task regularized linear regression due to its simplicity and good performance. It appears important to combine the training data across all subject groups for the most effective predictive models.  
\end{abstract}


\titlepgskip=-15pt

\maketitle

\section{Introduction}
\label{sec:introduction}
\emph{Alzheimer’s disease} (AD) is a chronic neurodegenerative disorder and a major health burden, with 152 million people expected to suffer from AD by 2050~\cite{Patterson2018world}. Pathophysiological changes in AD begin many years prior to clinical manifestations of disease, and the spectrum of AD spans from clinically asymptomatic to severely impaired~\cite{aisen2017path}. Because of this, there is an appreciation that AD should not only be viewed with discrete and defined clinical stages, but also as a multifaceted process moving along a continuum. Mild cognitive impairment (MCI) is an essential concept along this continuum, representing a transitional stage between healthy elderly individuals and AD~\cite{langa2014diagnosis}. Approximately 10\% to 20\% of MCI patients tend to progress to AD annually, whereas others will continue with cognitive decline or even revert to normal cognition (NC)~\cite{koepsell2012reversion}. Many treatment strategies have been proposed 
to decelerate AD, with limited success~\cite{cummings2020alzheimer}; one problem is that treatments are not administered within the correct time window along the AD continuum. Therefore, early prediction of disease progression 
is a crucial step towards better therapies, unburdening the health care system, and preventing adverse events caused by AD~\cite{de2013impact, rasmussen2019alzheimer}.

Cognitive test batteries have been developed to assess the cognitive decline of individuals. Two of the most commonly used standards are the Mini-Mental State Examination (MMSE)~\cite{folstein1975mini} and the Alzheimer’s Disease Assessment Scale-cognition subscale (ADAS-Cog)~\cite{mohs1997development}, which are important criteria for the clinical diagnosis of AD. In the evaluation of cognitive decline due to dementia, ADAS-Cog is considered to be more sensitive and reliable than MMSE~\cite{KAUFMAN2017105}. The ADAS-Cog is widely used to evaluate the cognitive state of patients with mild to advanced AD. The modified ADAS-Cog 13 contains 13 items for assessing cognitive dysfunction, with a total 
score of 0–85, with higher scores indicating greater dysfunction~\cite{skinner2012alzheimer}.

Due to the above-stated reasons  \emph{machine learning} (ML) for predicting ADAS-Cog scores, 
as opposed to the diagnosis (a recent review in~\cite{ansart2020predicting}), has been gaining research interest. For example, Utsumi et al.~\cite{utsumil2018personalized} found that the combination of two variants of personalized Gaussian process models can improve the accuracy of predicting future ADAS-Cog13 scores using a limited set of subjects with multimodal data, which included imaging biomarkers (MRI, diffusion tensor imaging), cerebrospinal fluid (CSF) biomarkers, demographics, genetics, and cognitive scores. Zhu et al.~\cite{zhu2016canonical} concluded that the canonical feature selection method had a significant effect on improving the performance of sparse multitask learning (MTL) to predict the clinical scores of ADAS-Cog. Prakash et al.~\cite{prakash2020quantitative} utilized multivariate regression techniques and determined that longitudinal prediction of AD progression is possible with multimodal data from the baseline, which included MRI, positron emission tomography (PET), CSF biomarkers, cognitive scores and APOE. Tsao et al.~\cite{tsao2017feature} found that leveraging hippocampal surface features together with multimodal data, which included sex, age, MRI, APOE, and baseline MMSE, might boost the prediction of cognitive scores, such as MMSE and clinical dementia rating (CDR). 
However, PET is far less commonly used and more expensive than MRI~\cite{wittenberg2019economic}, and obtaining CSF is highly invasive.

Accordingly, an interesting approach to biomarkers is the use of the only standard, T1-weighted MRI, to predict the progression rate of dementia. MRI enables 
the study of various noninvasive aspects of the human brain to detect biomarkers associated with AD~\cite{vemuri2010role}, and it is a widely available imaging modality. Indeed, several studies have investigated the association between cognitive scores and MRI biomarkers, such as gray matter volume~\cite{wang2010high,jie2016temporally} and cortical and subcortical volumes~\cite{cao2018L2, cao2019feature}. Lei et al.~\cite{lei2020deep} studied the relationship between MRI data and cognitive scores by introducing a framework that includes correntropy regularized joint learning and a deep polynomial network for feature construction, as well as ensemble learning based on support vector regression for the prediction of cognitive scores. Bhagwat et al.~\cite{bhagwat2019artificial} proposed an artificial neural network model for predicting cognitive scores from the cortical thickness and hippocampal subfield volumes. Jiang et al.~\cite{jiang2018correlation} proposed a novel MTL formulation that considers a correlation-aware sparse and low-rank constrained regularization in order to explore the relationship between the MRI features and cognitive scores. Huang et al.~\cite{huang2016longitudinal} presented a random forest (RF) with sparse regression and soft-split technique, which adopted probabilistic paths during the testing stage in RF to predict cognitive scores at multiple time points. Zhou et al.~\cite{zhou2013modeling} proposed two MTL formulations based on a temporal group Lasso regularizer and the convex fused sparse group Lasso, which utilize the common temporal patterns of MRI biomarkers to predict disease progression measured by the cognitive scores. To date, most existing progression models focus on predicting cognitive scores derived from the entire AD continuum, from healthy elderly to moderate AD, using a single model, e.g.,  ~\cite{stonnington2010predicting,zhou2011multi,yan2015cortical,lu2020predicting}, with the exception of Duchesne et al.~\cite{duchesne2009relating}, who studied the relationship between MRI and one-year MMSE changes in the MCI population. However, there is little evidence that this one-size-fits-all strategy would be optimal. Moreover, individuals at different stages of the continuum can be expected to regress differently (for example, most cognitively normal individuals are likely to be cognitively normal 
after three years, while most 
AD patients are expected to have regressed during that time period). Therefore, 
evaluating the prediction models using the whole-continuum data leads to results that are perhaps hard to interpret, and we argue that the prediction models should be evaluated while stratifying the subject population based on the baseline diagnosis.

The evaluation of the prediction models while stratifying for the baseline diagnosis leads to a question of whether the other diagnostic groups are still useful when training the predictive models. This subsequently leads to the consideration of MTL approaches to improve the generalization performance by simultaneously solving multiple learning tasks while exploiting commonalities and differences across tasks~\cite{jalali2013dirty}. One of the critical issues in MTL is to identify the essential relatedness between the tasks and to build learning models in order to obtain this task relatedness. MTL approaches with sparsity-inducing regularization have been studied to investigate the prediction of cognitive measures. For example, Tabarestani et al.~\cite{tabarestani2020distributed} applied $\ell$1-norm regularization to introduce sparsity among all features that could select a small subset of features to predict MMSE at six time points. Zhou et al.~\cite{zhou2013modeling} and Lei et al.~\cite{lei2017longitudinal} employed joint sparsity regularization ($\ell$2,1 norm) in order to share a common subset of features for all tasks simultaneously, where each task refers to AD progression prediction at a single time point. Wang et al.~\cite{wang2020modeling} formulated the progression of AD as a weakly supervised temporal multitask matrix regression framework that considers the prediction of cognitive scores at each time point as a regression task. However, MTL has not been studied in cases where different tasks correspond to cognitive score prediction of different diagnostic groups within the AD continuum. In addition, MTL has not been studied for adapting predictive modeling for differences in MRI acquisition. 

In this study, we explore whether MRI at the baseline can potentially predict changes in ADAS-Cog scores while stratifying the population based on the baseline diagnosis. We will compare single and multitask learning approaches for the task. We will also address multitask learning in the presence of differences in terms of MRI acquisition; in this case, MRIs have been acquired using two different magnetic field strengths (MFSs). We compare MTL to two heterogeneity reduction approaches, partial least squares (PLS) domain adaptation \cite{moradi2017predicting} and ComBat \cite{johnson2007adjusting}.

\section{Materials and methods}
\subsection{ADNI dataset}
Data used in the preparation of this article were obtained from the Alzheimer’s Disease
Neuroimaging Initiative (ADNI) database (adni.loni.usc.edu) \footnote{For up-to-date information, see \url{https://www.adni-info.org}.}. 
The ADNI was launched in 2003 as a public-private partnership, led by Principal Investigator Michael W. Weiner,
MD. The primary goal of ADNI has been to test whether serial magnetic resonance imaging
(MRI), positron emission tomography (PET), other biological markers, and clinical and
neuropsychological assessments can be combined to measure the progression of MCI and early AD. 


\subsection{Subjects and MRI}
The data in this study included baseline MRIs from 1376 ADNI subjects (430 NC, 662 MCI, 284 AD), aged 54 to 91 years old. Detailed characteristics of the subjects are presented in Table~\ref{Tab_1:Demographic}. For these subjects, the baseline MRI data were obtained with a T1-weighted MP-RAGE sequence at 1.5 T (with a 256 × 256 × 170 acquisition matrix and a voxel size of 1.25 × 1.25 × 1.2 $mm^{3}$) and 3.0 T (with a 256 × 256 × 170 acquisition matrix and a voxel size of 1.0 × 1.0 × 1.2 $mm^{3}$). Specifically, 808 subjects were from the ADNI-1 cohort with the MRI acquired at 1.5 T, and 571 subjects were from the ADNI-2 cohort with the MRI acquired at 3.0 T. As 
seen in Table \ref{Tab_1:Demographic}, the number of subjects with ADAS-Cog-13 scores decreased during the follow-up due to subject drop-out and missing data. The roster identification numbers (RIDs) of the subjects employed in this study are provided in the Supplementary Material. 
\begin{table*}[!t]
\footnotesize
\begin{center}
\caption{Demographic information of subjects at different time points. Under the ranges, the mean and standard deviation are provided.}
\setlength{\tabcolsep}{15.0pt} 
\renewcommand{\arraystretch}{1.28} 
\vspace{1mm}
\begin{tabular}{l c c c c c}

\hline
\hline
 Time point & Baseline diagnosis & No. of Subjects  & Age & Male/Female & ADAS-Cog \Tstrut  \\ 
  & & (1.5 T, 3.0 T) & & & \\ \cline{1-5}
 \hline
 \multirow{8}{*}{Baseline} & All  & 1376   & Range: [54-91]  & 759/617  & Range: [0-55] \Tstrut\\
          &          & (808, 568) & 74.05(7.10)   &  & 17.34(9.59) \\
          & NC  & 430   & Range: [56-90]         &214/216  & Range: [0-24] \\
          &         & (227, 203)  & 74.39(5.77)   &  & 9.38(4.27) \\
          & MCI & 662   & Range: [54-91]         & 394/268 & Range: [3-40] \\
          &        &  (394, 268)  & 73.43(7.60)   &  & 17.19(6.67) \\
          & AD  & 284  & Range: [55-91]         &  151/133& Range: [13-55] \\
          &       &    (187, 97) & 74.95(7.59)   &  & 29.77(8.03) \\
&  &  & & & \\
\hline
 \multirow{8}{*}{12-Months} & All & 1160  & Range: [55-91]  & 647/513 & Range: [0-73] \Tstrut   \\
          &      &   (711, 449)   & 74.24(7.04)   &  & 18.34(11.97) \\
          & NC & 347   & Range: [56-90]  & 177/170 & Range: [0-23] \\
          &     &   (206, 141)    & 74.89(5.61)   &  & 8.75(4.59) \\
          & MCI & 594   & Range: [55-91]  & 352/242 & Range: [0-73] \\
          &    &    (351, 243)    & 73.47(7.57)   &  & 18.17(8.89) \\
          & AD  & 219   & Range: [56-91]  & 118/101 & Range: [9-71] \\
          &   &    (154, 65)     & 75.30(7.39)   &  & 33.99(11.02) \\
          &  &  & & & \\
\hline

\multirow{8}{*}{24-Months} & All  & 1011    & Range: [55-91]  &559/452  & Range: [0-71] \Tstrut \\
          &  &     (617, 394)      & 73.98(6.97)   &  & 18.55(13.16) \\
          & NC  & 367   & Range: [56-90]         & 189/178  &Range: [0-26] \\
          &       &  (200, 167)   & 74.41(5.83)   &  & 9.06(4.86) \\
          & MCI & 501   & Range: [55-91]         & 297/204 &Range: [0-71] \\
          &      &  (296, 205)    & 73.30(7.46)   &  & 19.85(10.46) \\
          & AD & 143  & Range: [56-89]         & 73/70 & Range: [12-68] \\
          &     &    (121, 22)   & 75.25(7.63)   &  & 38.40(12.25) \\
&  &  & & & \\
\hline
\multirow{8}{*}{36-Months} & All  & 629        & Range: [55-90]  & 353/276 & Range: [0-74] \Tstrut \\
          &      &  (437, 192)    & 73.75(6.99)   &  & 17.09(11.96) \\
          & NC  & 208  & Range: [56-90]         & 104/104 & Range: [0-33] \\
          &       &  (182, 26)   & 75.30(5.46)   &  & 9.32(5.24) \\
          & MCI & 411  & Range: [55-87]         & 245/166 & Range: [0-74] \\
          &        &  (245, 166)  & 72.86(7.52)   &  & 20.49(12.09) \\
          & AD  & 10   & Range: [66-85]         & 4/6 & Range: [22-73] \\
          &     &    (10, 0)   & 77.95(5.51)   &  & 39.33(15.14) \\
&  &  & & & \\
       \hline
       
       \hline
    
\end{tabular}
\label{Tab_1:Demographic}
\end{center}
\vspace{0mm}
\end{table*}

\subsection{Image preprocessing}
The preprocessing of the T1-weighted images was performed using the fully automated CAT12 package running under MATLAB\footnotemark. T1-weighted images were first denoised by using adaptive nonlocal means filtering ~\cite{manjon2010adaptive}, then they were corrected for bias field inhomogeneities and segmented 
into gray matter (GM), white matter (WM), and cerebrospinal fluid (CSF) ~\cite{rajapakse1997statistical}. After segmentation, partial volume estimation (PVE) with a simplified mixed model with a maximum of two tissue types was performed, resulting in maps of the tissue type densities ~\cite{tohka2004fast}. Furthermore, the segmented images were spatially normalized by utilizing the high-dimensional DARTEL normalization algorithm into the standard MNI space ~\cite{ashburner2007fast}. This procedure resulted in spatially aligned maps for tissue fractions of WM and GM. We only utilized the GM images in this study. Finally, we averaged the gray matter density values according to the brain regions defined by the AAL atlas, resulting in 122 regional GM density values.   
\footnotetext{For more information, see \url{http://www.neuro.uni-jena.de/cat}.}

\subsection{ADAS-Cog score}

The ADAS-Cog was developed as an outcome measure to assess the severity of cognitive dysfunction in AD. We consider ADAS-Cog-13, which yields a measure of cognitive performance by combining the original tasks of the ADAS-Cog-11 ~\cite{rosen1984new} (subject-completed tests and observer-based assessments) as well as a test of delayed word recall and a number cancellation or maze task~\cite{mohs1997development}. The ADAS-Cog-13, later referred to as ADAS-Cog, scores are in the range of 0 to 85, with higher scores indicating more severe impairment ~\cite{kueper2018alzheimer}. The ADAS-Cog scores used in this study were acquired once a year as described in the ADNI General Procedures Manuals\footnote{ \url{http://adni.loni.usc.edu/wp-content/uploads/2010/09 \\ /ADNI\_ GeneralProceduresManual.pdf}}.

\subsection{Overview of the methods}
\begin{figure*}[!t]
\centering
\includegraphics[width=0.99\textwidth]{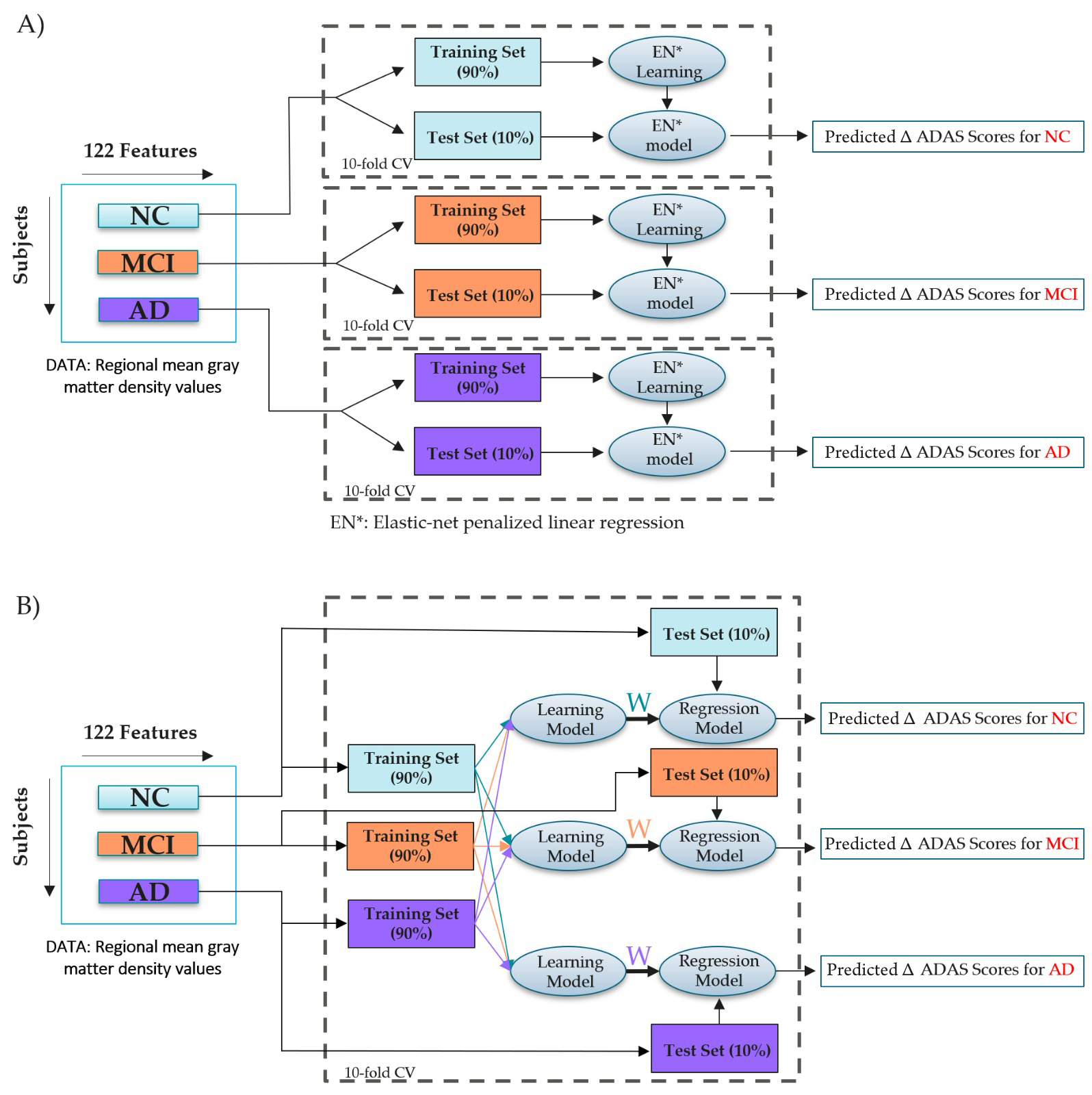}
\caption{The framework of two different learning models to predict the future change of ADAS-Cog scores. A) Single-task learning, B) Multitask learning.} 
\label{Fig_1:Single_Multi}
\vspace{0mm}
\end{figure*}

We aim to predict the change in the ADAS-Cog score ($\Delta_t ADAS_{i}$) for each subject $i$ and each time point ($t = 12, 24, 36$ months) based on regional, MRI-derived gray matter density values at the baseline ($t = 0$ months). The change in the ADAS-Cog score is defined as $\Delta_t ADAS_{i} =  ADAS_{i}(t) - ADAS_{i}(0)$, where $ADAS_{i}(t)$ is the ADAS score of subject $i$ at time $t$. We frame this as a single-task (Fig.~\ref{Fig_1:Single_Multi}(A)) or multitask (Fig.~\ref{Fig_1:Single_Multi}(B)) prediction problem, where different tasks correspond to the groups of different, but related, subjects (three diagnostic groups and/or two magnetic field strengths used to acquire the data). More formally, we build $S$ prediction models: 
\begin{equation}
\Delta_t ADAS_{i} = f_d(x_i),
\end{equation}
where $x_i$ denotes the MRI (122 regional gray matter density values) of subject $i$ at baseline and $f_d$ is the prediction model for subject group $d$. We consider the cases where $S = 3$, where the groups are determined based on the baseline diagnosis (NC, MCI, and AD), and $S = 6$, where the groups are determined based on the baseline diagnosis and MFS (NC at 1.5 T, NC at 3 T, MCI at 1.5 T, MCI at 3 T, AD at 1.5 T and AD at 3 T). multitask learning aimed at enhancing the precision of learning algorithms by jointly learning independent variables for multiple tasks. The learning approach works well, especially if these tasks have some commonalities, as we expect $f_d$ to have.

We compare multitask learning to harmonizing MRI acquired with different field strengths. We do this by adopting a recent domain adaptation technique ~\cite{moradi2017predicting} and using ComBat data harmonization ~\cite{fortin2017harmonization}. The algorithms compared in this work are summarized in Table~\ref{Tab_2:Sum_Methods}. Supplementary Figure 1 delineates the workflow of the applied methods for predicting the change of ADAS-Cog scores.

\begin{table*}[!t]
\footnotesize
\begin{center}
\caption{Summary of the compared algorithms. $\mathcal{L}$(W) and E(W) represent the empirical loss and the regularization term, respectively.}
\setlength{\tabcolsep}{12.5pt} 
\renewcommand{\arraystretch}{1.48} 
 \begin{tabular}{l c l  c  l }
\hline
\hline
Purpose &  &Method &   & Algorithm\\ \cline{1-5} 
 \hline

\multirow{9}{*}{MFS harmonization}& & ComBat &    & ComBat harmonization. \\

& &\multirow{2}{*}{ComBat$_{Age}$} &    & ComBat harmonization;  \\
& & &  & Age as a covariate. \\

& &\multirow{2}{*}{ComBat$_{Reg_{Age}}$} &     & Eliminate age as a confound  from  \\
& & &  & harmonized image data. \\
& &$PLS$ & & PLS based domain adaptation; MFSs as the response variable. \\

&&\multirow{2}{*}{$PLS_{Age}$} &     & PLS based domain adaptation; MFSs along with age  \\
& & &  & as the response variable. \\

&&\multirow{1}{*}{Dirty Model$_{6-Tasks}$} &      & 6-task Dirty learning model \\ 
 
& & & & \\ 
 \hline

\multirow{12}{*}{Learning models}& &\multirow{2}{*}{SEP-EN } &    & Elastic-net penalized linear regression; Training the model  \\ 
& & &  &  separately for each group. \\
&&\multirow{2}{*}{ALL-EN } &    & Elastic-net penalized linear regression; Training the model  \\ 
& & &  &  using all groups \\
&&\multirow{2}{*}{Least Lasso} &    & Multitask learning; $\mathcal{L}$(W) = Least Squares;   \\ 
& & &  & E(W) = $\rho_{1} ||W||_{1} + \rho_{L2}$. \\
&&\multirow{2}{*}{JFS} &    & Multitask learning; $\mathcal{L}$(W) = Least Squares;   \\  
& & &  & E(W) = $\lambda ||W||_{1,2}$. \\
&&\multirow{2}{*}{Dirty Model} &    & Multitask learning; $\mathcal{L}$(W) = Least Squares;   \\  
& & &  & E(W) = $\rho_{1} ||R||_{1,\infty} + \rho_{2} ||S||_{1}$, W = R+S.\\
&&\multirow{2}{*}{LRA} &    & Multitask learning; $\mathcal{L}$(W) = Least Squares;  \\
& & &  & E(W) = $\lambda ||W||_{*}$. \\
       \hline
       
       \hline
    
\end{tabular}
\label{Tab_2:Sum_Methods}
\end{center}
\vspace{0mm}
\end{table*}

\subsection{Penalized linear regression}
\label{Penalized_linear_regression}
In the simplest cases, we assume that either 1) $f_d$ are independent or 2) $f_d$ are 
equal and can proceed with single-task learning. As a single-task learning algorithm, we use least-squares linear regression with an elastic net penalty to predict the ADAS-Cog changes. 
More formally, the $\Delta$ADAS scores are predicted by solving the following linear regression problems:
\begin{equation}
\Delta_t ADAS_{i} = a_d^T x_i + b_d +\epsilon_{i},
\end{equation}
where $i$ refers to a subject, $a_d$ and $b_d$ are the model
parameters for the task $d$ and $\epsilon_{i}$ is the error term. Adding the elastic net penalty, the model is solved by minimizing the following elastic net cost function:
\begin{eqnarray}
& &      \frac{1}{2N_d}\sum_{i=1}^{N_d} (\Delta_t ADAS_{i} - b_d - x_i^T  a_d)^2 \\ \nonumber
    & +  & \lambda [(1 - \alpha) || a_d||_{2}^{2}/2 + \alpha|| a_d||_1], 
\end{eqnarray}
where $N_d$ is the number of training samples, $\lambda$ is the complexity parameter found by cross-validation, $\alpha \in [0,1]$ defines the compromise between ridge $||a||_{2}^2/2$ and lasso penalties $||a||_1$, and $||$.$||_{1}$ denotes the L1-norm. Here, we selected $\alpha = 0.5$ to give equal weights for the ridge and lasso penalties. We modeled each time point ($t = 12, 24, 36$ months) separately.

\subsection{Multitask linear regression }
Multitask regression incorporates the shared information among different tasks (task relatedness) into the regression model. To introduce the multitask regression approaches, we need to introduce additional notation. Let $X_d$ be an $N_d \times M$ matrix of the input MRI data at the baseline corresponding to task $d$, where $N_{d}$ is the number of subjects in a group $d$, and $M$ is the dimensionality of the feature space. We formulate the multitask learning as minimization of the penalized empirical loss: 
\begin{equation}
\min_{W} \mathcal{L}(W) + E(W),
\end{equation}
where $W \in \mathds{R}^{M \times d}$ is the weight matrix, which is estimated from the feature matrices $X_d$ and $\Delta$ ADAS-Cog scores in the training set, $\mathcal{L}(W)$ is the empirical loss on the training set, and $E(W)$ is the regularization term that encodes the relatedness of different tasks. Moreover, we use $w_d$ to denote the weights related to the group $d$, i.e., $W = [w_{1},\ldots, w_{S}]^T$. We then use the least squares loss as the empirical loss function for the regression tasks:
\begin{equation}
   \mathcal{L}(W_t) = \sum_{d \in \{AD, MCI,NC\}}  || w_{d,t}^{T} X_{d} + b_{d,t}{\bf 1} - y_{d,t}||^{2},
\end{equation}
where $\y_{d,t}$ is a shorthand of the $\Delta_t ADAS$ scores of $N_d$ subjects in group $d$ at the time point $t$ (12, 24, or 36 months), $b_{d,t}$ are the bias terms, and ${\bf 1}$ is an $M$-element vector of ones. Different regularization terms were used to determine different assumptions on task relatedness ~\cite{argyriou2008convex,ji2009accelerated,obozinski2010joint,zhou2011malsar}.
The regularization terms are described as in the following, where we drop the subindex $t$ as we separately consider the prediction at three time points.

\subsubsection{Multitask Lasso}

Multitask Lasso is a simple generalization of the elastic net penalty ~\cite{tibshirani1996regression} to multitask regression. The regularization term is defined as:

\begin{equation}
E(W) = \rho_{1} ||W||_{1} + \rho_{L2} ||W||_{F}^2,
\end{equation}
where $\rho_{1}$ is a regularization parameter for controlling the sparsity among all tasks and $\rho_{L2}$ is an optional regularization parameter that controls the $\ell$2-norm penalty.

\subsubsection{Joint feature selection}

Joint feature selection is used to constrain all models to share a common set of features ~\cite{argyriou2007multi, argyriou2008convex}. This goal is achieved by setting $E(W)$ and minimizing the following $\ell$2,1-norm regularized learning ~\cite{liu2012multi,nie2010efficient}:

\begin{equation}
E(W) = \rho_{1} ||W||_{2,1} + \rho_{L2} ||W||_{F}^2,
\end{equation}
where $||W||_{1,2} = \sum_{D=1}^3 ||W_{D}||_{2}$, with $W_{D} \in \mathds{R}^{1 \times D}$, is the group sparse penalty, $\rho_{1}$ is a regularization parameter for controlling the sparsity among all tasks and $\rho_{L2}$ is an optional regularization parameter that controls the $\ell$2-norm penalty. 

\subsubsection{Dirty model for multitask learning}

The Dirty model estimates a superposition of two sets of parameters and regularizes them differently ~\cite{jalali2010dirty}. In more detail, each $w_d$ is written as a sum $w_d = s_d + r_d$, where the corresponding matrices $S$ and $R$ are encouraged to have elementwise sparsity and block-structured row sparsity. This leads to the minimization problem

\begin{eqnarray}
& \min_{S,R} & \sum_{d} ||(s_{d} + r_d)^{T} X_{d} + b_d{\bf 1} - y_{d,t}||^{2} \\ \nonumber
& &  +  \rho_{1} ||R||_{1,\infty} + \rho_{2} ||S||_{1} ,  
\end{eqnarray}
where $||R||_{1,\infty} = \sum_j \max_d|r_{d_j}|$. The final output is $W = \hat{S} + \hat{R}$, where $\hat{S},\hat{R}$ are the minimizers of (8).

\subsubsection{Low rank assumption} 

The task commonalities can be utilized to constrain the prediction models from different tasks to share a low-dimensional subspace, i.e., constrain $W$ to be of low rank ~\cite{he2016novel}. Since rank optimization is, in general, NP-hard, the rank function ~\cite{fazel2003matrix} is replaced by the trace norm  ~\cite{ji2009accelerated,obozinski2010joint}, which 
is given by the sum of the singular values of $W$: $||W||_{*} = \sum_{d}\sigma_{d}(W)$, and the regularization term is

\begin{equation}
E(W) = \rho_{1} ||W||_{*}, 
\end{equation}
where the regularization parameter $\rho_{1}$ controls the rank of $W$.

\subsection{Magnetic field strength harmonization} 

To correct for differences in features caused by imaging at two MFSs in ADNI1 and ADNI2, we studied three different techniques: 1) PLS-based domain adaptation introduced in ~\cite{moradi2017predicting}, 2) ComBat harmonization originating from genetics \cite{johnson2007adjusting}, which has become widely used in brain imaging \cite{fortin2017harmonization,fortin2018harmonization}, and 3) multitask learning.

\subsubsection{Partial least squares domain adaptation}

We next briefly explain the PLS-based domain adaptation method introduced in \cite{moradi2017predicting} for correcting the site dependency of cortical thickness measurements in 
predicting the severity of autism spectrum disorder. PLS is a linear feature transformation method for modeling relations between two sets of observed variables. The idea of PLS-based domain adaptation is to project the input features into a new (lower-dimensional) space, which is a product of two orthogonal subspaces: a subspace that is dependent on MFSs and its (orthogonal) complement. This can be achieved by applying a PLS algorithm that ensures the orthogonality of the resulting latent features so that the response variable is a codification of the scanner MFS and the predictor variables are the original features. However, as demonstrated in \cite{moradi2017predicting}, this works the best, and different brain regions are separately corrected. Therefore, \cite{moradi2017predicting} suggested a two-stage strategy where in the first stage a PLS-based domain adaptation is performed for each brain region separately, then for each brain region the prediction task is performed. These predictions are then combined in the stacking framework. We applied an AAL-atlas to decompose the gray matter density values into 122 distinct regions. For the prediction $\Delta ADAS$ for each brain region, we utilized support vector regression (with a radial basis function kernel), as suggested in \cite{moradi2017predicting}. Finally, the per-region predictions were combined using the elastic-net penalized linear regression described in Section~\ref{Penalized_linear_regression}.

\subsubsection{ComBat harmonization}

We applied ComBat harmonization to reduce any potential biases induced by different MFSs. This method was initially proposed to correct for the site effects in genomics~\cite{johnson2007adjusting}. Later, ComBat was applied to correct for site effects in imaging applications, including  diffusion tensor imaging data~\cite{fortin2017harmonization}, cortical thickness  ~\cite{fortin2018harmonization}, positron emission tomography ~\cite{orlhac2018postreconstruction} and functional connectivity  ~\cite{yu2018statistical}. In this study, ComBat utilizes a multivariate linear mixed-effects regression to model MFS-adapted feature measurements. Let $x_{ij}$ denote a regional gray matter density value for subject $j$ with MRI acquired at MFS $i \in \{1, 2\}$ (where 1 refers to $1.5 T$ and 2 refers to $3 T$). Then, $x_{ij}$ can be written as:

\begin{equation}
x_{ij} = \alpha + C_{ij}\beta + \gamma_{i} +\delta_{i}\epsilon_{ij},
\end{equation}
where $\alpha$ is the overall GM density, $C_{ij}$ is a design matrix for the covariates of interest (age), and $\beta$ is the regression coefficient corresponding to the covariate $C$. The terms $\gamma_{i}$ represent the location parameter effect of MFS $i$, $\delta_{i}$ describes the multiplicative effect of MFS $i$, and $\epsilon_{ij}$ is an error term from a normal distribution with a zero mean ~\cite{fortin2017harmonization}. The ComBat-harmonized GM densities (MFSs-adjusted) are then defined as:

\begin{equation}
x_{ij}^{ComBat} = \frac{x_{ij}-\widehat{\alpha} - C_{ij}\widehat{\beta} - \gamma_{i}^{*}}{\delta_{i}^{*}} + \widehat{\alpha} + C_{ij}\widehat{\beta},
\end{equation}
in which $\widehat{\alpha}$ and $\widehat{\beta}$ are estimators of parameters $\alpha$ and $\beta$, and $\gamma_{i}^{*}$ and $\delta_{i}^{*}$ are the empirical Bayes estimators of $\gamma_{i}$ and $\delta_{i}$. To correct the difference among MFSs, we considered two strategies: 1) using ComBat without adjusting any biological covariates (i.e., setting $C=0$) and 2) using ComBat while adjusting the age as a biological covariate.

\subsection{Implementation and evaluation}

To evaluate the performance of the models, we utilized repeated, nested 10-fold cross-validation (CV). We used two evaluation metrics: (1) $R$ is the correlation coefficient between the predicted and observed $\Delta$ ADAS-Cog scores, averaged over ten repeats of the CV; and (2) MAE is the mean absolute error between the observed and predicted $\Delta$ ADAS-Cog values, averaged over the subjects and ten CV repeats. We then used ten repeats of the 10-fold CV and averaged the metrics to reduce the random variation due to the sampling of subjects to different folds. We computed $95\%$ confidence intervals for cross-validated, averaged correlations $R$ and MAEs using a bootstrap method ~\cite{good2006permutation,lewis2018t1}. Confidence intervals represent an approximation of the overall performance of the prediction model, and the specific bootstrap method used is adapted to be used in repeated CV (see \cite{lewis2018t1} for details). 

The values of the hyperparameters were selected in the inner CV loop, and predictions of the $\Delta$ ADAS-Cog scores were evaluated in the outer CV loop to avoid the problem of training on the testing data. The distribution of the ADAS-Cog score changes in each of the CV folds was similar since we used stratified cross-validation folds\footnote{The stratified cross-validation code is available\\ at \url{https://github.com/jussitohka/general_matlab}.} ~\cite{huttunen2012meg}. 

The implementation of the elastic-net penalized linear regression model was performed by using the glmnet library\footnote{The glmnet library is available at \\ \url{https://web.stanford.edu/~hastie/glmnet_matlab}.}~\cite{qian2013glmnet}. The optimal value of the regularization parameter ($\lambda$) was selected in the 10-fold inner CV loop by minimizing the mean squared error (MSE). The implementation of multitask learning techniques was performed using the MALSAR package running in MATLAB ~\cite{zhou2011malsar}. The functions implemented in MALSAR have many parameters to tune. Since fine-tuning all parameters with a grid search was impractical, we only considered $\rho_{1}$ (the regularization parameter for controlling the sparsity among all tasks) and $\rho_{2}$ (an optional regularization parameter that controls the $\ell$2-norm penalty) as the most crucial parameters for the grid search. The $\rho_{1}$ parameter was selected among the candidate set $\{10^{-3}, 10^{-2.5}, \ldots, 10^{2}, 2 \cdot 10^{2}, 2.5 \cdot 10^{2}, \ldots, 5 \cdot 10^{2})\}$, where the parameter $\rho_{2}$ was chosen among the candidate set $\{10^{-3}, 10^{-2.5}, \ldots, 10^{2}, 2 \cdot 10^{2}  2.5 \cdot 10^{2}, \ldots, 10 \cdot 10^{2})\}$ by minimizing the root-mean-square error (rmse). For the tuning parameters, default values were used for the optional optimization parameters (starting points, termination criterion, endurance, and a maximum number of repetitions).

The implementation of ComBat was performed using a publicly available MATLAB package\footnote{The ComBat harmonization package is available at \url{https://github.com/Jfortin1/ComBatHarmonization}.}. PLS was performed by the PLSREGRESS function in MATLAB with a constant number of components for all groups of diagnoses at each time point. The number of components for PLS and PCA was selected at each time point by initial experiments between the candidate set $\{5, 10, 15, 17, 20, 25\}$. 

The PLS-based domain adaption was performed as instructed in \cite{moradi2017predicting}. The implementation of SVR, required by the PLS domain adaptation, was performed using LIBSVM\footnote{\url{https://www.csie.ntu.edu.tw/~cjlin/libsvm/}.} ~\cite{chang2011libsvm}. The SVR model parameters were set to their default values $ ( C = 1$, $\nu = 0.5$, $\lambda = 1/K$, where $K$ refers to the dimensionality of the feature space$)$. The implementation of elastic-net penalized linear regression, required by the PLS domain adaptation, was performed as described above.

\begin{table*}[!t]
\scriptsize
\begin{center}
\caption{Comparison of single and multitask learning on predicting the change in ADAS-Cog. The methods are given in Table~\ref{Tab_2:Sum_Methods}. SEP-EN refers to the single-task method trained for each diagnostic group separately. ALL-EN refers to the single-task method trained with the data from all diagnostic groups. $R$ is the cross-validated correlation between the actual and predicted $\Delta$ ADAS-Cog scores averaged over 10 CV runs and $MAE$ is the mean absolute error averaged over 10 CV runs. Values in parentheses give the bootstrapped 95\% confidence intervals. The asterisk (*) implies that the validation result is not trustworthy due to the low number of samples.}
\setlength{\tabcolsep}{6.9pt} 
\renewcommand{\arraystretch}{1.5} 
 \begin{tabular}{l c c c c c c c c}
\hline
\hline
 & \multicolumn{2}{c}{NC} & \multicolumn{2}{c}{MCI} & \multicolumn{2}{c}{AD} & \multicolumn{2}{c}{ALL}   \\ \cline{1-9}
 \hline

& \hspace{0.01in} $R$ \hspace{0.01in} & \hspace{0.01in}MAE\hspace{0.01in} & \hspace{0.01in} $R$ \hspace{0.01in} & \hspace{0.01in}MAE\hspace{0.01in} & \hspace{0.01in} $R$ \hspace{0.01in} & \hspace{0.01in}MAE\hspace{0.01in} & \hspace{0.01in} $R$ \hspace{0.01in} & \hspace{0.01in}MAE\hspace{0.01in}\\ 

\multicolumn{2}{l}{SEP-EN }  &  &  & &  &  & &  \\ 
\multirow{2}{*}{$\Delta$ ADAS-12} & $0.09$ &$3.07$  & $0.19$ & $3.84$&$0.24$  &$4.86$ & $0.38$&$3.81$ \\
   &(0.00 $to$ 0.19)  &(2.79 $to$ 3.32)  &(0.11 $to$ 0.27) &(3.58 $to$ 4.17)  &(0.12 $to$ 0.35)  &(4.33 $to$ 5.49) &(0.32 $to$ 0.43)&(3.60 $to$ 4.01)  \\ 
   
\multirow{2}{*}{$\Delta$ ADAS-24} & $0.09$ & $3.16$ & $0.36$&$4.74$ &$0.40$&$ 6.46$ &$0.55$ &$4.43$  \\ 

&(0.00 $to$ 0.19)  &(2.93 $to$ 3.41)  &(0.29 $to$ 0.44) &(4.37 $to$ 5.13)  &(0.23 $to$ 0.52)  &(5.74 $to$ 7.37) &(0.49 $to$ 0.60)&(4.18 $to$ 4.65)  \\  
\multirow{2}{*}{$\Delta$ ADAS-36} & $0.04$ & $3.25$ & $0.37$ &$5.95$ & $-0.28^{*}$ &  $7.06^{*}$ &$0.43$ & $5.07$\\

&(-0.02 $to$ 0.11)  &(2.90 $to$ 3.69)  &(0.29 $to$ 0.45) &(5.43 $to$ 6.50)  &(-0.67 $to$ 0.33)  &(3.81 $to$ 10.35) &(0.37 $to$ 0.50)&(4.73 $to$ 5.52)  \\  
     &  &  &  & &  &  & &  \\

\multicolumn{2}{l}{ALL-EN }   &  &  & &  &  & &  \\
\multirow{2}{*}{$\Delta$ ADAS-12} & $0.12$ & $3.23$  & $0.22$ & $3.84$ & $0.22$ & $4.74$  & $0.32$&$3.83$  \\

&(0.03 $to$ 0.23)  &(2.96 $to$ 3.49)  &(0.15 $to$ 0.30) &(3.56 $to$ 4.12)  &(0.06 $to$ 0.34)  &(4.10 $to$ 5.39) &(0.27 $to$ 0.38)&(3.63 $to$ 4.04)  \\  

\multirow{2}{*}{$\Delta$ ADAS-24}& $0.17$ & $3.59$ & $0.41$ & $4.65$ & $0.26$ & $6.55$ & $0.48$ & $4.54$   \\ 

&(0.07 $to$ 0.28)  &(3.36 $to$ 3.89)  &(0.34 $to$ 0.48) &(4.28 $to$ 5.05)  &(0.11 $to$ 0.40)  &(5.46 $to$ 7.66) &(0.44 $to$ 0.53)&(4.29 $to$4.82)  \\  

\multirow{2}{*}{$\Delta$ ADAS-36} & $0.19$ & $3.86$ &  $0.39$ & $5.83$ & $0.71^{*}$ & $5.07^{*}$ & $0.42$ & $5.17$  \\

&(0.09 $to$ 0.30)  &(3.47 $to$ 4.27)  &(0.32 $to$ 0.46) &(5.28 $to$ 6.39)  &(-0.07 $to$ 0.90)  &(2.19 $to$ 8.58) &(0.35 $to$ 0.48)&(4.81 $to$ 5.56)  \\  

     &  &  &  & &  &  & &  \\   

\multicolumn{2}{l}{Least Lasso}& & & & & & &    \\

\multirow{2}{*}{$\Delta$ ADAS-12} & $0.10$ & $3.07$ & $0.17$ & $3.87$ & $0.26$ & $5.01$ & $0.33$ & $3.85$ \\

&(0.01 $to$ 0.21)  &(2.81 $to$ 3.33)  &(0.10 $to$ 0.26) &(3.56 $to$ 4.21)  &(0.15 $to$ 0.37)  &(4.35 $to$ 5.60) &(0.26 $to$ 0.40)&(3.63 $to$ 4.07)  \\  

\multirow{2}{*}{$\Delta$ ADAS-24}&$0.09$&$3.15$&$0.37$&$5.10$&$0.29$&$7.36$&$0.47$&$4.70$   \\
&(0.02 $to$ 0.22)  &(2.92 $to$ 3.42)  &(0.29 $to$ 0.45) &(4.69 $to$ 5.46)  &(0.11 $to$ 0.43)  &(6.30 $to$ 8.59) &(0.40 $to$ 0.52)&(4.41 $to$ 4.97)  \\  

\multirow{2}{*}{$\Delta$ ADAS-36}&$0.15$&$3.24$&$0.37$&$6.46$&$-0.17^{*}$&$8.16^{*}$&$0.38$&$5.34$   \\
&(0.07 $to$ 0.30)  &(2.88 $to$ 3.65)  &(0.30 $to$ 0.46) &(5.72 $to$ 6.95)  &(-0.53 $to$ 0.14)  &(3.85 $to$ 13.47) &(0.31 $to$ 0.45)&(4.92 $to$ 5.80)  \\  

     &  &  &  & &  &  & &  \\   

\multicolumn{2}{l}{JFS}&  &  &  & &  &  &  \\

\multirow{2}{*}{$\Delta$ ADAS-12}&$0.11$&$3.07$&$0.20$&$3.86$&$0.25$&$4.99$&$0.34$&$3.84$ \\

&(0.01 $to$ 0.21)  &(2.81 $to$ 3.34)  &(0.13 $to$ 0.29) &(3.56 $to$ 4.14)  &(0.13 $to$ 0.37)  &(4.31 $to$ 5.59) &(0.28 $to$ 0.40)&(3.62 $to$ 4.05)  \\

\multirow{2}{*}{$\Delta$ ADAS-24}&$0.10$&$3.17$&$0.37$&$5.03$&$0.30$&$7.24$&$0.48$&$4.67$    \\ 

&(0.02 $to$ 0.22)  &(2.93 $to$ 3.39)  &(0.29 $to$ 0.45) &(4.63 $to$ 5.47)  &(0.14 $to$ 0.42)  &(6.23 $to$ 8.35) &(0.42 $to$ 0.53)&(4.40 $to$ 4.96)  \\

\multirow{2}{*}{$\Delta$ ADAS-36}&$0.09$&$3.25$&$0.38$&$6.33$&$0.38^{*}$&$6.77^{*}$&$0.40$&$5.33$   \\
&(0.01 $to$ 0.21)  &(2.98 $to$ 3.68)  &(0.30 $to$ 0.46) &(5.76 $to$ 6.91)  &(-0.13 $to$ 0.79)  &(3.62 $to$ 10.35) &(0.32 $to$ 0.46)&(4.91 $to$ 5.80)  \\  

     &  &  &  & &  &  & &  \\   

\multicolumn{2}{l}{Dirty Model}&&&&&&&  \\

\multirow{2}{*}{$\Delta$ ADAS-12}&$0.09$&$3.09$&$0.21$&$3.84$&$0.27$&$4.96$&$0.34$&$3.82$  \\

&(0.01 $to$ 0.18)  &(2.81 $to$ 3.35)  &(0.13 $to$ 0.28) &(3.55 $to$ 4.15)  &(0.15 $to$ 0.38)  &(4.28 $to$ 5.61) &(0.28 $to$ 0.40)&(3.62 $to$ 4.06)  \\  

\multirow{2}{*}{$\Delta$ ADAS-24}&$0.08$&$3.18$&$0.37$&$5.06$&$0.27$&$7.18$&$0.47$&$4.70$    \\ 

&(-0.01 $to$ 0.16)  &(2.96 $to$ 3.46)  &(0.29 $to$ 0.45) &(4.63 $to$ 5.48)  &(0.12 $to$ 0.41)  &(6.20 $to$ 8.53) &(0.41 $to$ 0.53)&(4.43 $to$ 4.99)  \\  

\multirow{2}{*}{$\Delta$ ADAS-36}&$0.18$&$3.27$&$0.39$&$6.29$&$0.38^{*}$&$5.87^{*}$&$0.41$&$5.27$   \\

&(0.05 $to$ 0.30)  &(2.92 $to$ 3.70)  &(0.31 $to$ 0.45) &(5.71 $to$ 6.94)  &(-0.30 $to$ 0.76)  &(3.32 $to$ 8.55) &(0.33 $to$ 0.48)&(4.90 $to$ 5.76)  \\  

     &  &  &  & &  &  & &  \\   

\multicolumn{2}{l}{LRA}&  &  &  & &  &  &   \\

\multirow{2}{*}{$\Delta$ ADAS-12}&$0.09$&$3.08$&$0.21$&$3.84$&$0.25$&$4.96$&$0.33$&$3.82$  \\

&(-0.01 $to$ 0.20)  &(2.79 $to$ 3.36)  &(0.13 $to$ 0.29) &(3.54 $to$ 4.16)  &(0.12 $to$ 0.37)  &(4.34 $to$ 5.66) &(0.27 $to$ 0.40)&(3.61 $to$ 4.04)  \\

\multirow{2}{*}{$\Delta$ ADAS-24}&$0.13$&$3.16$&$0.39$&$5.00$&$0.31$&$7.11$&$0.49$&$4.65$    \\ 

&(0.03 $to$ 0.23)  &(2.94 $to$ 3.41)  &(0.31 $to$ 0.46) &(4.64 $to$ 5.42)  &(0.16 $to$ 0.45)  &(6.11 $to$ 8.32) &(0.43 $to$ 0.55)&(4.38 $to$ 4.94)  \\  

\multirow{2}{*}{$\Delta$ ADAS-36} & $0.01$ & $3.48$ & $0.34$ & $6.62$ & $0.28^{*}$ & $6.66^{*}$ & $0.35$ & $5.56$   \\
&(-0.08 $to$ 0.10)  &(3.13 $to$ 3.94)  &(0.27 $to$ 0.43) &(6.00 $to$ 7.19)  &(-0.44 $to$ 0.82)  &(3.82 $to$ 9.12) &(0.28 $to$ 0.43)&(5.16 $to$ 6.03)  \\
  &  &  &  & &  &  & &  \\
  
       \hline
       
       \hline
\end{tabular}
\label{Tab_3:Single_Multi_Comparison}
\end{center}
\end{table*}

\begin{figure*}[!t]
\centering
\includegraphics[width=0.99\textwidth]{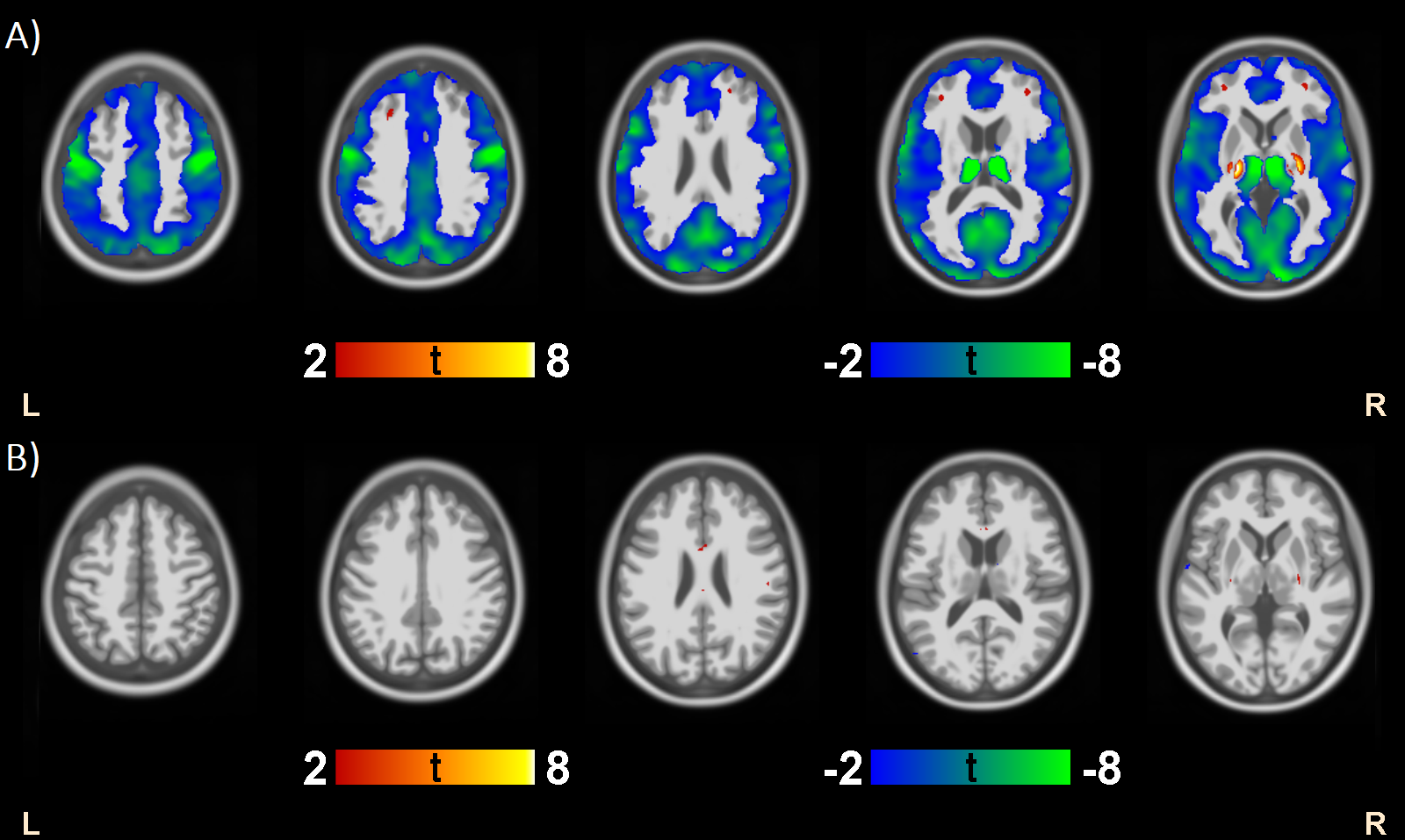}
\caption{t-scores of voxel-wise differences in the gray matter density in cognitively normal subjects due to differences in magnetic field strength. The t-score values have a threshold at $\left|t\right| > 2$, corresponding to uncorrected $p < 0.05$. Panel (A) before using the ComBat harmonization and (B) after using the ComBat harmonization approach. The differences in the gray matter density were strong in panel (A), but ComBat harmonization was successful in removing the contribution of MFS at the group level.}
\label{Fig_5:t_test_Before_ComBat}
\vspace{3mm}
\end{figure*}

\section{Results}

\label{results}

\subsection{Prediction performance of single- and multitask learning}

We evaluated the performance of single- and multitask learning approaches by predicting the future change in ADAS-Cog scores using baseline MRI features. The experimental results presented in this subsection ignore the variation in the magnetic field strengths to acquire the MRIs. 

Table~\ref{Tab_3:Single_Multi_Comparison} indicates the comparison results between different multitask learning methods based on the least-squares loss function, including multitask Lasso (least lasso), joint feature selection (JFS), dirty model (least dirty), and trace-norm regularization (least trace), with two single-task learning strategies (SEP-EN and ALL-EN) based on elastic-net penalized linear regression (EN). Note that the low number of subjects in the AD group at 36 months (only ten subjects) cannot provide a reliable validation of predictive models; however, the results for this group and time point are shown in Table~\ref{Tab_3:Single_Multi_Comparison}. As shown in Table~\ref{Tab_3:Single_Multi_Comparison}, the average correlation coefficients between the predicted and actual $\Delta$ ADAS-Cog scores were positive for all baseline diagnoses and time points. This demonstrates that MRI-based predictive models were able to predict the disease progression. A comparison of two single-task learning strategies, SEP-EN and ALL-EN, indicated that ALL-EN, which utilized all diagnostic groups for training, performed better in the NC and MCI groups. 
For instance, $R$ of the NC subjects increased from 0.09 to 0.12 at 12 months, from 0.09 to 0.17 at 24 months, and from 0.04 to 0.19 at 36 months. Moreover, the ALL-EN prediction model achieved the best performance among all methods in terms of a correlation score for NC and MCI groups (e.g., $R$ for MCI group at 12, 24, and 36 months were 0.22, 0.41, and 0.39). However, $R$ of the other methods were typically within the 95\% confidence intervals of $R$ of ALL-EN, indicating that the improvement was not large. All multitask learning algorithms performed highly similarly to ALL-EN. However, especially in the predictions concerning the AD group, these performed slightly better than ALL-EN: For instance, $R$ of the AD subjects increased from 0.22 to 0.27 at 12 months and from 0.26 to 0.27 at 24 months when comparing the Dirty Model and ALL-EN. SEP-EN was clearly the method of choice for predicting 24-month changes in the ADAS-Cog in AD patients, with a $R$ value of 0.40 compared to the maximum of 0.31 of others. 
Supplementary Figures 2-3 provide scatter plots of the observed vs. predicted change in ADAS scores in the CV run with the median correlation ($R$). These scatter plots imply that the $\Delta$ ADAS-Cog scores with very high values for all time points were the most difficult to predict, since the number of individuals with observed ADAS-Cog score changes over 20 was small (e.g., the numbers of MCI subjects at 12, 24, and 36 months were 2, 8, and 19, respectively). In summary, these results support the notion that auxiliary data were useful in predicting the ADAS-Cog change in the NC and MCI groups, but in the AD group. Complex multitask learning algorithms did not demonstrate benefits over simpler single-task learning methods.

Interestingly, in the majority of methods, $R$ scores typically increased with the length of the follow-up (e.g., $R$ in $\Delta$ ADAS-24 was higher than $R$ of $\Delta$ ADAS-12). The potential reason for this higher correlation is that as the changes in ADAS-Cog become more prominent, they are easier to predict based on MRI.
 
Table~\ref{Tab_3:Single_Multi_Comparison} the ALL column lists the evaluation results while agglomerating all subject groups in the validation. For all methods, the $R$ values were higher when all the subjects were combined than 
stratified based on the baseline diagnosis. Since the prediction models were the same in both cases, this inflation in the prediction performance can be seen as artificial and one to avoid. It is likely a product of the interaction between the heterogeneity of subject groups and the particular evaluation measure (correlation) that scales according to this heterogeneity.

\subsection{Prediction performance while accounting for differences in MFS}

This subsection focuses on the situation associated with heterogeneity reduction between data from two MFSs and explores the prediction models by including MFS correction approaches to reduce unwanted variance across features. In addition, based on the results of the previous subsection, we selected ALL-EN as a single-task learning method and a Dirty model as a multitask learning method for further analysis.

\begin{table*}[!t]
\scriptsize
\centering
\caption{Comparison of the predictive model performance with and without MFS harmonization. ALL-EN refers to the baseline method without any harmonization. $ComBat_{ALL-EN}$ ($PLS_{ALL-EN}$) refers to Combat (PLS) harmonization followed by ALL-EN. $ComBat_{Dirty Model}$( $PLS_{Dirty Model}$) refers to Combat (PLS) harmonization followed by Dirty Model.$Dirty Model_{6-Tasks}$ ($LRA_{6-Tasks}$) refers to the 6-task learning with Dirty Model (LRA). The asterisk (*) implies that the validation result is not trustworthy due to the low number of samples.}
\setlength{\tabcolsep}{5.5pt} 
\renewcommand{\arraystretch}{1.3} 
 \begin{tabular}{l c c c c c c c c}
\hline
\hline
 & \multicolumn{2}{c}{NC} & \multicolumn{2}{c}{MCI} & \multicolumn{2}{c}{AD} & \multicolumn{2}{c}{ALL}   \\ \cline{1-9}
 \hline

& \hspace{0.01in} $R$ \hspace{0.01in} & \hspace{0.01in}MAE\hspace{0.01in} & \hspace{0.01in} $R$ \hspace{0.01in} & \hspace{0.01in}MAE\hspace{0.01in} & \hspace{0.01in} $R$ \hspace{0.01in} & \hspace{0.01in}MAE\hspace{0.01in} & \hspace{0.01in} $R$ \hspace{0.01in} & \hspace{0.01in}MAE\hspace{0.01in}\\

\multicolumn{2}{l}{ALL-EN }   &  &  & &  &  & &  \\
\multirow{2}{*}{$\Delta$ ADAS-12} & $0.12$ & $3.23$  & $0.22$ & $3.84$ & $0.22$ & $4.74$  & $0.32$&$3.83$  \\

&(0.03 $to$ 0.23)  &(2.96 $to$ 3.49)  &(0.15 $to$ 0.30) &(3.56 $to$ 4.12)  &(0.06 $to$ 0.34)  &(4.10 $to$ 5.39) &(0.27 $to$ 0.38)&(3.63 $to$ 4.04)  \\  

\multirow{2}{*}{$\Delta$ ADAS-24}& $0.17$ & $3.59$ & $0.41$ & $4.65$ & $0.26$ & $6.55$ & $0.48$ & $4.54$   \\

&(0.07 $to$ 0.28)  &(3.36 $to$ 3.89)  &(0.34 $to$ 0.48) &(4.28 $to$ 5.05)  &(0.11 $to$ 0.40)  &(5.46 $to$ 7.66) &(0.44 $to$ 0.53)&(4.29 $to$4.82)  \\  

\multirow{2}{*}{$\Delta$ ADAS-36} & $0.19$ & $3.86$ &  $0.39$ & $5.83$ & $0.71^{*}$ & $5.07^{*}$ & $0.42$ & $5.17$  \\ 

&(0.09 $to$ 0.30)  &(3.47 $to$ 4.27)  &(0.32 $to$ 0.46) &(5.28 $to$ 6.39)  &(-0.07 $to$ 0.90)  &(2.19 $to$ 8.58) &(0.35 $to$ 0.48)&(4.81 $to$ 5.56)  \\  

 &  &  &  & &  &  & &  \\
$ComBat_{ALL-EN}$  &  &  & &  &  & & &  \\

\multirow{2}{*}{$\Delta$ ADAS-12} & $0.10$ &$3.23$  & $0.21$ & $3.86$&$0.20$  &$4.75$ & $0.31$&$3.84$ \\

&(-0.01 $to$ 0.19)  &(2.98 $to$ 3.51)  &(0.13 $to$ 0.29) &(3.57 $to$ 4.17)  &(0.05 $to$ 0.34)  &(4.06 $to$ 5.40)  &(0.26 $to$ 0.36)&(3.63 $to$ 4.06)  \\ 
     
\multirow{2}{*}{$\Delta$ ADAS-24} & $0.11 $ & $3.67 $ & $0.38 $&$4.76 $ &$0.27$  &$6.64$  &$  0.45 $ &$ 4.64 $  \\

&(0.02 $to$ 0.21)  &(3.41 $to$ 3.95)  &(0.29 $to$ 0.45) &(4.40 $to$ 5.16)  &(0.12 $to$ 0.42)  &(5.69 $to$ 7.78) &(0.40 $to$ 0.49)&(4.39 $to$ 4.91)  \\ 

\multirow{2}{*}{$\Delta$ ADAS-36} & $0.18 $ & $3.85 $ & $0.37 $ &$5.94 $ & $0.70^{*}$ &$4.99^{*}$ 
 &$0.41 $ & $5.22 $\\

&(0.06 $to$ 0.30)  &(3.42 $to$ 4.23)  &(0.29 $to$ 0.44) &(5.40 $to$ 6.49)  &(0.03 $to$ 0.91)  &(2.60 $to$ 7.87)  &(0.34 $to$ 0.47)&(4.85 $to$ 5.61)  \\ 

     &  &  &  & &  &  & &  \\

$PLS_{ALL-EN}$&   &  &  & &  &  & &  \\
\multirow{2}{*}{$\Delta$ ADAS-12} & $0.11$ & $3.27 $  & $0.23$ & $3.83 $ & $0.30$ & $4.67 $  & $0.34 $&$3.83 $  \\

&(0.01 $to$ 0.21)  &(3.00 $to$ 3.56)  &(0.16 $to$ 0.32) &(3.56 $to$ 4.12)  &(0.16 $to$ 0.43)  &(4.05 $to$ 5.27)&(0.27 $to$ 0.39)&(3.62 $to$ 4.05) \\

\multirow{2}{*}{$\Delta$ ADAS-24} & $0.16$ & $3.85$ & $0.40$ & $4.82$ & $0.28$ & $6.58$ & $0.48$ & $4.70$   \\ 

&(0.06 $to$ 0.25)  &(3.53 $to$ 4.09)  &(0.34 $to$ 0.47) &(4.48 $to$ 5.21)  &(0.13 $to$ 0.43)  &(5.55 $to$ 7.59)&(0.43 $to$ 0.52)&(4.45 $to$ 4.95) \\

\multirow{2}{*}{$\Delta$ ADAS-36} & $0.22$ & $3.99$ &  $0.36$ & $5.96 $ & $0.77^{*} $ & $5.15^{*} $ & $0.40 $ & $5.26 $  \\

&(0.10 $to$ 0.34)  &(3.56 $to$ 4.35)  &(0.30 $to$ 0.44) &(5.40 $to$ 6.49)  &(0.34 $to$ 0.91)  &(2.47 $to$ 7.97)&(0.34 $to$ 0.46)&(4.92 $to$ 5.65) \\

     &  &  &  & &  &  & &  \\ 
\multicolumn{2}{l}{$ComBat_{Dirty Model}$}  &  &  & &  &  & &  \\

\multirow{2}{*}{$\Delta$ ADAS-12} & $0.05$ &$3.11$  & $0.19$ & $3.86$&$0.27$  &$4.92$ & $0.33$&$3.84$ \\

&(-0.04 $to$ 0.14)  &(2.83 $to$ 3.36)  &(0.11 $to$ 0.27) &(3.59 $to$ 4.19)  &(0.13 $to$ 0.38)  &(4.30 $to$ 5.58) &(0.27 $to$ 0.38)&(3.63 $to$ 4.07)  \\

\multirow{2}{*}{$\Delta$ ADAS-24} & $0.01 $ & $3.24 $ & $0.34 $&$5.12 $ &$0.27$&$ 7.72 $ &$  0.42 $ &$ 4.81 $  \\

&(-0.08 $to$ 0.07)  &(3.02 $to$ 3.52)  &(0.26 $to$ 0.42) &(4.71 $to$ 5.52)  &(0.12 $to$ 0.40)  &(6.56 $to$ 8.91) &(0.36 $to$ 0.48)&(4.52 $to$ 5.13)  \\

\multirow{2}{*}{$\Delta$ ADAS-36} & $0.16$ & $3.30 $ & $0.38 $ &$6.33 $ & $0.58^{*}$ &  $5.63^{*}$ &$0.40 $ & $5.31 $\\

&(0.02 $to$ 0.26)  &(2.93 $to$ 3.71)  &(0.30 $to$ 0.45) &(5.72 $to$ 6.94)  &(0.01 $to$ 0.89)  &(3.45 $to$ 7.25) &(0.33 $to$ 0.47)&(4.88 $to$ 5.74)  \\

     &  &  &  & &  &  & &  \\

\multicolumn{2}{l}{$PLS_{Dirty Model}$}   &  &  & &  &  & &  \\
\multirow{2}{*}{$\Delta$ ADAS-12} & $0.09 $ & $3.11 $  & $0.18$ & $3.91 $ & $0.24 $ & $5.20 $  & $0.30 $&$3.93 $  \\

&(-0.01 $to$ 0.20)  &(2.86 $to$ 3.40)  &(0.12 $to$ 0.26) &(3.64 $to$ 4.26)  &(0.13 $to$ 0.35)  &(4.60 $to$ 5.94) &(0.24 $to$ 0.36)&(3.70 $to$ 4.15)  \\

\multirow{2}{*}{$\Delta$ ADAS-24} & $0.15$ & $3.20$ & $0.39$ & $4.99$ & $0.27$ & $7.11$ & $0.48$ & $4.62$   \\

&(0.02 $to$ 0.23)  &(2.95 $to$ 3.44)  &(0.31 $to$ 0.47) &(4.59 $to$ 5.45)  &(0.10 $to$ 0.43)  &(5.97 $to$ 8.30) &(0.43 $to$ 0.54)&(4.35 $to$ 4.93)  \\

\multirow{2}{*}{$\Delta$ ADAS-36} & $0.08$ & $3.34$ &  $0.38$ & $6.41 $ & $0.62^{*} $ & $7.23^{*} $ & $0.39 $ & $5.40 $  \\

&(-0.04 $to$ 0.18)  &(2.97 $to$ 3.73)  &(0.31 $to$ 0.46) &(5.79 $to$ 7.05)  &(-0.42 $to$ 0.90)  &(3.97 $to$ 10.52) &(0.32 $to$ 0.46)&(4.98 $to$ 5.88)  \\

     &  &  &  & &  &  & &  \\  
\multicolumn{2}{l}{$Dirty Model_{6-Tasks}$}   &  &  & &  &  & &  \\
\multirow{2}{*}{$\Delta$ ADAS-12} &$0.09$ & $3.11$ & $0.19$ & $3.91$ & $0.14$ & $4.98$ & $0.34$ & $3.88$  \\

&(0.00 $to$ 0.18) & (2.87 $to$ 3.40) & (0.12 $to$ 0.25) & (3.66 $to$ 4.23) & (0.02 $to$ 0.26) & (4.40 $to$ 5.62) & (0.29 $to$ 0.39) & (3.68 $to$ 4.09)    \\

\multirow{2}{*}{$\Delta$ ADAS-24} &$-0.005$ & $3.76$ & $0.35$ & $4.93$ & $0.35$ & $6.59$ & $0.50$ & $4.74$  \\

&(-0.09 $to$ 0.08) & (3.50 $to$ 4.06) & (0.27 $to$ 0.43) & (4.56 $to$ 5.28) & (0.20 $to$ 0.48) & (5.73 $to$ 7.45) & (0.44 $to$ 0.55) & (4.49 $to$ 5.02)  \\

\multirow{2}{*}{$\Delta$ ADAS-36} & $0.14$ & $3.36$ & $0.37$ & $6.00$ & $0.41^{*}$ & $5.34^{*}$ & $0.44$ & $5.12$  \\

&(0.00 $to$ 0.25) & (2.98 $to$ 3.78) & (0.29 $to$ 0.45) & (5.51 $to$ 6.55) & (-0.08 $to$ 0.78) & (2.43 $to$ 8.72) & (0.38 $to$ 0.50) & (4.76 $to$ 5.52)  \\

     &  &  &  & &  &  & &  \\  
\multicolumn{2}{l}{$LRA_{6-Tasks}$}   &  &  & &  &  & &  \\
\multirow{2}{*}{$\Delta$ ADAS-12} &$0.02$ & $3.15$ & $0.18$ & $3.92$ & $0.16$ & $5.02$ & $0.33$ & $3.90$  \\

&(-0.06 $to$ 0.11) & (2.88 $to$ 3.43) & (0.11 $to$ 0.25) & (3.66 $to$ 4.23) & (0.04 $to$ 0.28) & (4.44 $to$ 5.64) & (0.27 $to$ 0.38) & (3.70 $to$ 4.12)  \\

\multirow{2}{*}{$\Delta$ ADAS-24} & $0.02$ & $3.60$ & $0.35$ & $4.82$ & $0.25$ & $6.77$ & $0.49$ & $4.65$   \\

&
(-0.06 $to$ 0.10) & (3.32 $to$ 3.87) & (0.27 $to$ 0.42) & (4.46 $to$ 5.23) & (0.10 $to$ 0.41) & (5.83 $to$ 7.69) & (0.44 $to$ 0.54) & (4.40 $to$ 4.94)  \\

\multirow{2}{*}{$\Delta$ ADAS-36} & $0.03$ & $3.43$ & $0.40$ & $5.98$ & $-0.40^{*}$ & $6.90^{*}$ & $0.44$ & $5.15$  \\

&(-0.07 $to$ 0.14) & (3.02 $to$ 3.83) & (0.31 $to$ 0.48) & (5.50 $to$ 6.53) & (-0.74 $to$ -0.03) & (3.57 $to$ 10.78) & (0.37 $to$ 0.52) & (4.77 $to$ 5.54)  \\

     &  &  &  & &  &  & &  \\
       \hline
       
       \hline
    
\end{tabular}
\label{Tab_5:ComBat_PLS_ALLD_LR}
\end{table*}

To demonstrate the differences between GM density values of images acquired at 1.5 T and 3.0 T, we applied a standard voxel-based morphometry approach to compare GM densities of NC subjects acquired at 1.5 T and 3.0 T. Voxelwise t-statistics in Fig. \ref{Fig_5:t_test_Before_ComBat}(A) demonstrate considerable differences in the GM density values between 1.5 T and 3.0 T. We repeated the analysis after using the ComBat harmonization method. Fig. \ref{Fig_5:t_test_Before_ComBat}(B) delineates that, at the group level, the ComBat harmonization performed exceptionally well in removing nuisance variability associated with two different MFSs.

\begin{figure*}[!t]
\begin{center}
\includegraphics[width=0.99\textwidth]{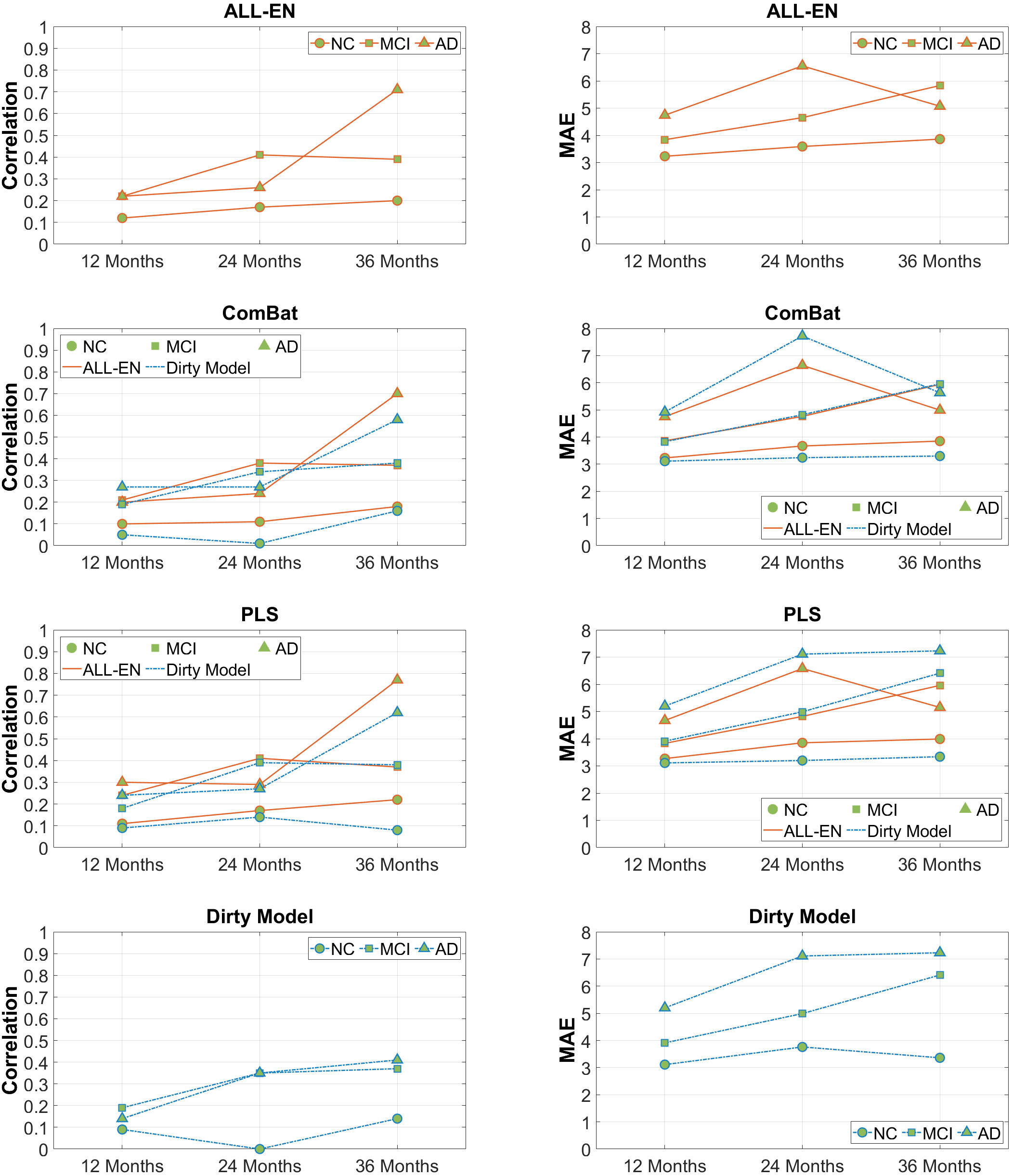}
\end{center}
\caption{Comparison of single and multitask learning with MFS correction (ComBat and PLS), without MFS correction (ALL-EN), and 6-task Dirty model. Combat and PLS panels give the results with both the ALL-EN and 3-task Dirty model.}
\label{Fig_6:Comparison_ComBat_PLS_WO_NEW}
\vspace{0mm}
\end{figure*}

\begin{table*}[!t]
\scriptsize
\begin{center}
\caption{Comparison of ComBat, PLS with Age as a covariate. In $ComBat_{Age}$ Age was used as covariate to preserve its effect while removing the variability associated with MFS. In $ComBat_{Reg_{Age}}$, age was regressed out from MRI after the MFS correction with  $ComBat_{Age}$. In $PLS_{age}$, Age and MFS were used as a response variables. The asterisk (*) implies that the validation result is not trustworthy due to the low number of samples.}
\setlength{\tabcolsep}{7.0pt} 
\renewcommand{\arraystretch}{1.7} 
 \begin{tabular}{l c c c c c c c c}
\hline
\hline
 & \multicolumn{2}{c}{NC} & \multicolumn{2}{c}{MCI} & \multicolumn{2}{c}{AD} & \multicolumn{2}{c}{ALL}   \\ \cline{1-9}
 \hline

& \hspace{0.01in} $R$ \hspace{0.01in} & \hspace{0.01in}MAE\hspace{0.01in} & \hspace{0.01in} $R$ \hspace{0.01in} & \hspace{0.01in}MAE\hspace{0.01in} & \hspace{0.01in} $R$ \hspace{0.01in} & \hspace{0.01in}MAE\hspace{0.01in} & \hspace{0.01in} $R$ \hspace{0.01in} & \hspace{0.01in}MAE\hspace{0.01in}\\

\multicolumn{5}{l}{$ComBat_{Age}$}  &  &  & &  \\ 

\multirow{2}{*}{$\Delta$ ADAS-12} & $0.10$ &$3.23$  & $0.20$ & $3.89$&$0.19$  &$4.75$ & $0.31$&$3.87$ \\

&(0.01 $to$ 0.21)  &(2.97 $to$ 3.50)  &(0.13 $to$ 0.28)&(3.60 $to$ 4.18)  &(0.05 $to$ 0.34)  &(4.08 $to$ 5.39) &(0.26 $to$ 0.37)&(3.66 $to$ 4.07)  \\ 
 
\multirow{2}{*}{$\Delta$ ADAS-24} & $0.08$ &$3.79$  & $0.36$ & $4.83$&$0.24$  &$6.77$ & $0.44$&$4.73$ \\

&(-0.01 $to$ 0.17)  &(3.54 $to$ 4.08)  &(0.29 $to$ 0.44)&(4.48 $to$ 5.20)  &(0.09 $to$ 0.38)  &(5.66 $to$ 7.90) &(0.39 $to$ 0.48)&(4.47 $to$ 5.01)  \\

\multirow{2}{*}{$\Delta$ ADAS-36} & $0.22$ &$3.81$  & $0.36$ & $6.00$&$0.65^{*}$ & $5.43^{*}$ & $0.41$&$5.25$ \\

&(0.10 $to$ 0.33)  &(3.44 $to$ 4.24)  &(0.29 $to$ 0.44)&(5.52 $to$ 6.51)  &(-0.12 $to$ 0.90)  &(2.45 $to$ 8.75) &(0.35 $to$ 0.47)&(4.89 $to$ 5.66)  \\

     &  &  &  & &  &  & &  \\
     

\multicolumn{5}{l}{$ComBat_{Reg_{Age}}$}   &  &  & &  \\ 

\multirow{2}{*}{$\Delta$ ADAS-12} & $0.11$ &$3.18$  & $0.19$ & $3.91$&$0.24$  &$4.75$ & $0.31$&$3.85$ \\

&(0.02 $to$ 0.21)  &(2.94 $to$ 3.47)  &(0.11 $to$ 0.27)&(3.61 $to$ 4.20)  &(0.07 $to$ 0.39)  &(4.12 $to$ 5.42) &(0.26 $to$ 0.37)&(3.63 $to$ 4.07)  \\ 
   
\multirow{2}{*}{$\Delta$ ADAS-24} & $0.11$ &$3.73$  & $0.39$ & $4.95$&$0.37$  &$6.54$ & $0.47$&$4.75$ \\

&(0.02 $to$ 0.20)  &(3.51 $to$ 4.02)  &(0.30 $to$ 0.45)&(4.63 $to$ 5.34)  &(0.21 $to$ 0.52)  &(5.50 $to$ 7.55) &(0.42 $to$ 0.52)&(4.49 $to$ 5.01)  \\

\multirow{2}{*}{$\Delta$ ADAS-36} & $0.16$ &$3.80$  & $0.38$ & $6.07$&$0.34^{*}$  &$5.57^{*}$ & $0.40$&$5.32$ \\

&(0.08 $to$ 0.33)  &(3.47 $to$ 4.27)  &(0.31 $to$ 0.46)&(5.40 $to$ 6.42)  &(0.15 $to$ 0.83)  &(2.26 $to$ 8.56) &(0.35 $to$ 0.47)&(4.81 $to$ 5.63)  \\

     &  &  &  & &  &  & &  \\

\multicolumn{5}{l}{$PLS_{Age}$}   &  &  & &  \\ 

\multirow{2}{*}{$\Delta$ ADAS-12} & $0.12$ &$3.27$  & $0.25$ & $3.79$&$0.31$  &$4.66$ & $0.35$&$3.80$ \\

&(0.03 $to$ 0.22)  &(3.02 $to$ 3.54)  &(0.18 $to$ 0.34)&(3.52 $to$ 4.11)  &(0.20 $to$ 0.42)  &(4.08 $to$ 5.23) &(0.29 $to$ 0.41)&(3.60 $to$ 4.03)  \\ 
   
\multirow{2}{*}{$\Delta$ ADAS-24} & $0.16$ &$3.80$  & $0.44$ & $4.78$&$0.34$  &$6.41$ & $0.51$&$4.84$ \\

&(0.06 $to$ 0.26)  &(3.48 $to$ 4.06)  &(0.38 $to$ 0.50)&(4.44 $to$ 5.10)  &(0.19 $to$ 0.49)  &(5.41 $to$ 7.38) &(0.46 $to$ 0.54)&(4.39 $to$ 4.88)  \\

\multirow{2}{*}{$\Delta$ ADAS-36} & $0.22$ &$3.92$  & $0.38$ & $5.92$&$0.75^{*}$  &$5.16^{*}$ & $0.41$&$5.27$ \\

&(0.08 $to$ 0.33)  &(3.60 $to$ 4.36)  &(0.31 $to$ 0.45)&(5.38 $to$ 6.48)  &(0.17 $to$ 0.92)  &(2.88 $to$ 8.39) &(0.35 $to$ 0.47)&(4.91 $to$ 5.63)  \\

     &  &  &  & &  &  & &  \\

       \hline
       
       \hline
    
\end{tabular}
\label{Tab_7:Biological_Covariate}
\end{center}
\vspace{3mm}
\end{table*}

We adopted two strategies to harmonize the MRI data for MFS differences and studied whether harmonization can improve the performance of ADAS-Cog prediction. First, we performed the ComBat approach on 122 regional GM density measurements and then used ALL-EN to predict $\Delta$ ADAS-Cog scores. Second, we applied the PLS-based domain adaptation method, as described in Section 2.8.1. Table \ref{Tab_5:ComBat_PLS_ALLD_LR} presents the $R$ and MAE scores. Combined with ALL-EN, PLS-based domain adaptation methods performed slightly better than the ComBat method in terms of the average correlation. For example, in the PLS approach, $R$ for the NC, MCI, and AD groups at 24 months were 0.16, 0.40, and 0.28, respectively. In the ComBat approach, $R$ for NC, MCI, and AD groups at 24 months were 0.11, 0.38, and 0.24, respectively.

Table~\ref{Tab_5:ComBat_PLS_ALLD_LR} delineates that the performance of the Dirty model was similar to or worse than the performance of ALL-EN after the correction for the MFS differences. 
For example, the prediction performance of the PLS method for the AD group at 12 months, when the Dirty model substituted the ALL-EN, the average $R$ score dropped from 0.30 to 0.24. Table~\ref{Tab_5:ComBat_PLS_ALLD_LR} also shows the results of 6-task  learning approaches for MFS adaptation. These methods were performed on par with other correction approaches, but failed to consistently improve the prediction of the baseline model (ALL-EN).  Fig.~\ref{Fig_6:Comparison_ComBat_PLS_WO_NEW} illustrates the performance comparison between single and multitask learning strategies before and after utilizing correction approaches. The performance comparison shows that ComBat did not improve the prediction performance at the individual level, although it worked well at the group level, as demonstrated in Fig.~\ref{Fig_5:t_test_Before_ComBat}. Fig.~\ref{Fig_6:Comparison_ComBat_PLS_WO_NEW} indicates that the PLS domain adaptation based on the single-task learning model performed consistently better than the other methods. Moreover, combining multitask learning with ComBat slightly improved the performance in the AD group.

\subsection{Age as a covariate}

\begin{figure*}[!t]
\centering
\includegraphics[width=0.99\textwidth]{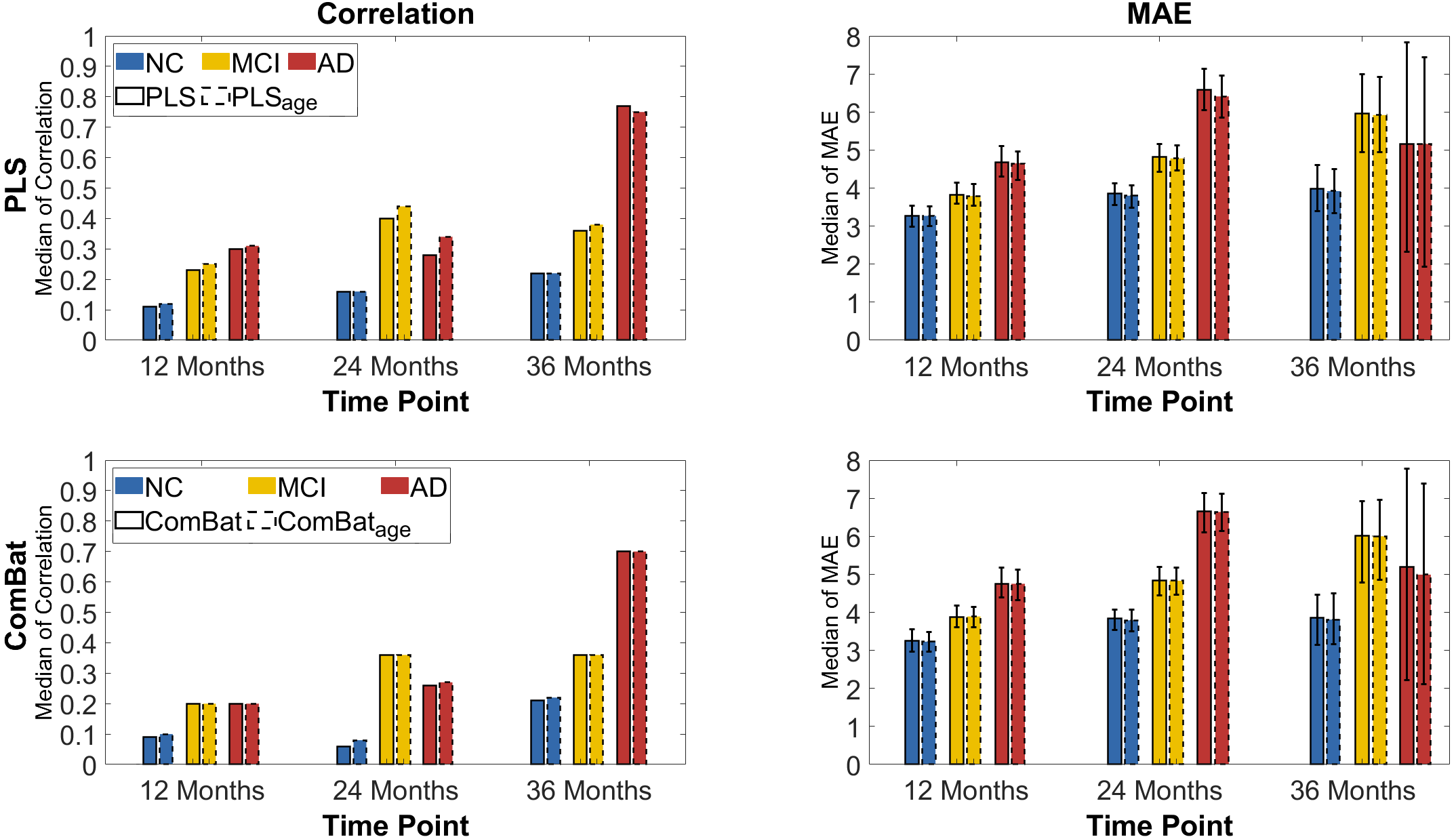}
\caption{Comparison of the accuracy of PLS and ComBat harmonization with age (dashed line) and without age (solid line) as a biological covariate. The figure provides the median MAE and R among 10 CV runs along with the $95\%$ confidence intervals obtained by the bootstrap method.}
\label{Fig_14:Compar_PLS_AGE}
\vspace{0mm}
\end{figure*}

In the MRI-based predictive modeling of AD, age plays an essential role; for example, regressing age out of MRI has been shown to improve MCI-to-AD conversion prediction \cite{moradi2015machine}. 
Therefore, we studied whether removing or preserving age as a biological variable can improve the ADAS-Cog prediction, focusing on the ALL-EN model. We considered different data harmonization methods: (1) we applied ComBat by considering age as a covariate to preserve its effect while removing the variability associated with MFS ($ComBat_{Age}$) and (2) additionally regressed age out of the MRI data before estimating the predictive model ($ComBat_{Reg_{Age}}$). (3) We applied PLS-based domain adaptation using age and MFS as response variables in PLS, effectively removing the effects of both MFS and age ($PLS_{Age}$). The prediction performances are listed in Table \ref{Tab_7:Biological_Covariate}. The comparison between $ComBat$ and $ComBat_{Age}$ (see Tables \ref{Tab_5:ComBat_PLS_ALLD_LR} and \ref{Tab_7:Biological_Covariate}) shows that the retaining age as a covariate did not improve $R$ or MAE. For instance, $R$ at 24 months dropped from 0.11 to 0.08 for NC, 0.38 to 0.36 for MCI, and 0.27 to 0.24 for AD. However, $ComBat_{Reg_{Age}}$ slightly improved the prediction performance for the AD group at 12 and 24 months: $R$ for the AD group at 12 and 24 months increased from 0.19 to 0.24 and from 0.24 to 0.37, respectively. $PLS_{Age}$ slightly improved the performance for each diagnosis group at the time points of 12 and 24 months compared to PLS without age as a covariate (see Table \ref{Tab_5:ComBat_PLS_ALLD_LR}). For example, $R$ increased from 0.40 to 0.44 for MCI and 0.28 to 0.34 for AD at 24 months. $PLS_{Age}$ featured slightly increased $R$ values for all diagnosis groups at all time points compared to Combat-based approaches, e.g., $R$ improved from 0.08 to 0.16 for NC, from 0.36 to 0.44 for MCI, and from 0.24 to 0.34 for AD at 24 months. The scatter plots are provided in Supplementary Figure 4. 
In addition, we illustrate the cross-validated accuracy of ComBat and PLS with and without considering age as a covariate in Fig. \ref{Fig_14:Compar_PLS_AGE}.

\section{Discussion}
We predicted the changes in the ADAS-Cog scores ($\Delta$ ADAS-Cog) in three distinct subject groups (NC, MCI, and AD) based on MRI for up to 36 months. We explored this problem by comparing various formulations of single- and multitask learning algorithms 
and scrutinizing whether multitask learning can help to cope with differences in the MRI data caused by different MFSs. 
MTL models aim to enhance generalization performance by utilizing relatedness among various tasks; here, predicting $\Delta$ ADAS-Cog in different subject groups.
We predicted $\Delta$ ADAS-Cog scores from regional GM density values by single-task learning via elastic net penalized linear regression as a baseline learning method. The single-task learning was applied based on two distinct strategies: 1) training the model by pooling together the data (ALL-EN) across the diagnostic groups, 
and 2) training a separate model for each diagnostic group (SEP-EN). We compared single-task models to multitask learning approaches, where we treated the prediction of cognitive scores in different baseline diagnoses as separate tasks.  

The experiments revealed a positive correlation between observed and predicted 
$\Delta$ ADAS-Cog scores in all diagnostic groups at all time points. This indicates that MRI has predictive value for changes in ADAS-Cog scores across all subject groups. 
As shown in Table~\ref{Tab_3:Single_Multi_Comparison}), the SEP-EN and MTL methods performed similarly; however, the ALL-EN method performed slightly better than the other methods 
regarding the average correlation score. In addition, a comparison of average correlation scores obtained from two single-task learning strategies (SEP-EN vs. ALL-EN) showed that simultaneous prediction in all diagnostic groups was beneficial for predicting disease progression in the NC and MCI groups. More complex multitask learning approaches were unable to provide benefits over single-task learning in our experiments. 

Considering different MFS used in ADNI1 and ADNI2 cohorts, we studied 1) whether correcting for this difference affects the ADAS-Cog prediction and 2) whether multitask learning would be useful for such a correction. We used two heterogeneity reduction approaches, typically applied for correcting for the site differences: PLS-based domain adaptation~\cite{moradi2017predicting} and ComBat~\cite{johnson2007adjusting}. Correcting the MRI with the help of PLS-based domain adaptation marginally improved the ADAS-Cog change prediction, but the improvement was typically not statistically significant, as seen by comparing the confidence intervals in Table \ref{Tab_5:ComBat_PLS_ALLD_LR}. Multitask learning with six tasks corresponding to the three baseline diagnoses and two MFSs of MRI did not bring any improvements over single-task learning.  

We investigated the role of age as a covariate in the prediction models. The evaluation demonstrated that the accuracy of predicted $\Delta$ ADAS-Cog scores improved by regressing out the age from the MRI data. This agrees with previous studies, indicating that age has a significant effect on the accuracy of cognitive score prediction~\cite{kueper2018alzheimer,stonnington2010predicting}. 

Several studies have analyzed the role of ADAS-Cog scores in the evaluation of AD, as well as the relationship between ADAS-Cog and MRI (\cite{stonnington2010predicting,zhou2011multi,zhou2013modeling,yan2015cortical,huang2016longitudinal,jiang2018correlation,wang2019multi,zandifar2020mri}). For instance, Wang et al.~\cite{wang2019multi} proposed a multitask exclusive relationship learning model to automatically capture the intrinsic relationship among tasks at different time points for estimating clinical measures based on longitudinal imaging data. Yan et al.~\cite{yan2015cortical} proposed a new group-sparse multitask regression model for predicting ADAS, MMSE, and RAVLT cognitive scores at the baseline using cortical thickness measurements. Duchesne et al.~\cite{duchesne2009relating} applied a linear regression model to predict one-year MMSE changes using baseline MRI features and revealed that the baseline MRI features moderately predict one-year MMSE changes in the general MCI population. However, to the best of our knowledge, this work is the first to stratify the estimation of the prediction performance between diagnostic groups and utilize the relatedness between the diagnostic groups to boost the prediction performance. In addition, we demonstrated the necessity of stratifying subjects based on a baseline diagnosis to evaluate the predictive modeling of the change in ADAS-Cog.

\section{Conclusion}

We explored single and multitask learning to predict the changes in ADAS-Cog scores based on T1-weighted anatomical MRI. We stratified the subjects based on their baseline diagnoses and evaluated the prediction performances in each group.  
Our results indicated a positive relationship between the predicted and observed ADAS-Cog score changes in each diagnostic group, suggesting that standard T1-weighted MRI has a predictive value for evaluating the cognitive decline in the AD continuum. We further studied whether correction of the differences in MFS of MRI would improve the ADAS-Cog score prediction. The PLS-based domain adaptation slightly improved the prediction performance, but the improvement was marginal. In summary, this study demonstrated that ADAS-Cog changes could be, to some extent, predicted based on anatomical MRI. Based on this study, the recommended method for learning the predictive models is ALL-EN, due to its simplicity and good performance.

\section*{Acknowledgments}

This study received funding from the Academy of Finland project 316258, "Predictive Brain Image Analysis", to Jussi Tohka.

 Data collection and sharing for this project were funded by the Alzheimer's Disease Neuroimaging Initiative (ADNI) (National Institutes of Health Grant U01 AG024904) and DOD ADNI (Department of Defense award number W81XWH-12-2-0012). ADNI is funded by the National Institute on Aging, the National Institute of Biomedical Imaging and Bioengineering, and through generous contributions from the following: AbbVie, Alzheimer's Association; Alzheimer's Drug Discovery Foundation; Araclon Biotech; BioClinica, Inc.; Biogen; Bristol-Myers Squibb Company; CereSpir, Inc.; Cogstate; Eisai Inc.; Elan Pharmaceuticals, Inc.; Eli Lilly and Company; EuroImmun; F. Hoffmann-La Roche Ltd and its affiliated company Genentech, Inc.; Fujirebio; GE Healthcare; IXICO Ltd.; Janssen Alzheimer Immunotherapy Research \& Development, LLC.; Johnson \& Johnson Pharmaceutical Research \& Development LLC.; Lumosity; Lundbeck; Merck \& Co., Inc.; Meso Scale Diagnostics, LLC.; NeuroRx Research; Neurotrack Technologies; Novartis Pharmaceuticals Corporation; Pfizer Inc.; Piramal Imaging; Servier; Takeda Pharmaceutical Company; and Transition Therapeutics. The Canadian Institutes of Health Research is providing funds to support ADNI clinical sites in Canada. Private sector contributions are facilitated by the Foundation for the National Institutes of Health (www.fnih.org). The grantee organization is the Northern California Institute for Research and Education, and the study is coordinated by the Alzheimer's Therapeutic Research Institute at the University of Southern California. ADNI data are disseminated by the Laboratory for Neuro Imaging at the University of Southern California.

\section*{Availability of Materials and methods }

The source codes of the single and multitask algorithms as well as MRI pre-processing are available at \url{https://github.com/vandadim/ADAS_MRI}. The dataset used in this paper was obtained from the Alzheimer's Disease Neuroimaging Initiative (ADNI) and is available at \url{http://adni.loni.usc.edu/data-samples/access-data/}.  The roster identification numbers (RIDs) of the subjects employed in this study are provided as a supplementary csv-file. 

\subsection*{Author contributions}
VI: Methodology, Software, Formal analysis, Writing - Original Draft, and Visualization. MP: Writing - Original Draft, and Visualization. MZ: Software, Data Curation, and Writing - Review and Editing. JT: Conceptualization, Methodology, Writing - Original Draft, Supervision, and Funding acquisition.

\bibliographystyle{IEEEtran}
\bibliography{sample.bib}

\begin{thebibliography}{10}
\providecommand{\url}[1]{#1}
\csname url@samestyle\endcsname
\providecommand{\newblock}{\relax}
\providecommand{\bibinfo}[2]{#2}
\providecommand{\BIBentrySTDinterwordspacing}{\spaceskip=0pt\relax}
\providecommand{\BIBentryALTinterwordstretchfactor}{4}
\providecommand{\BIBentryALTinterwordspacing}{\spaceskip=\fontdimen2\font plus
\BIBentryALTinterwordstretchfactor\fontdimen3\font minus
  \fontdimen4\font\relax}
\providecommand{\BIBforeignlanguage}[2]{{%
\expandafter\ifx\csname l@#1\endcsname\relax
\typeout{** WARNING: IEEEtran.bst: No hyphenation pattern has been}%
\typeout{** loaded for the language `#1'. Using the pattern for}%
\typeout{** the default language instead.}%
\else
\language=\csname l@#1\endcsname
\fi
#2}}
\providecommand{\BIBdecl}{\relax}
\BIBdecl

\bibitem{Patterson2018world}
C.~Patterson, \emph{World Alzheimer Report 2018: The state of the art of
  dementia research: New frontiers}.\hskip 1em plus 0.5em minus 0.4em\relax
  Alzheimer's Disease International, 2018.

\bibitem{aisen2017path}
P.~S. Aisen, J.~Cummings, C.~R. Jack, J.~C. Morris, R.~Sperling,
  L.~Fr{\"o}lich, R.~W. Jones, S.~A. Dowsett, B.~R. Matthews, J.~Raskin
  \emph{et~al.}, ``On the path to 2025: understanding the alzheimer’s disease
  continuum,'' \emph{Alzheimer's research \& therapy}, vol.~9, no.~1, pp.
  1--10, 2017.

\bibitem{langa2014diagnosis}
K.~M. Langa and D.~A. Levine, ``The diagnosis and management of mild cognitive
  impairment: a clinical review,'' \emph{Jama}, vol. 312, no.~23, pp.
  2551--2561, 2014.

\bibitem{koepsell2012reversion}
T.~D. Koepsell and S.~E. Monsell, ``Reversion from mild cognitive impairment to
  normal or near-normal cognition: risk factors and prognosis,''
  \emph{Neurology}, vol.~79, no.~15, pp. 1591--1598, 2012.

\bibitem{cummings2020alzheimer}
J.~Cummings, G.~Lee, A.~Ritter, M.~Sabbagh, and K.~Zhong, ``Alzheimer's disease
  drug development pipeline: 2020,'' \emph{Alzheimer's \& Dementia:
  Translational Research \& Clinical Interventions}, vol.~6, no.~1, p. e12050,
  2020.

\bibitem{de2013impact}
M.~E. de~Vugt and F.~R. Verhey, ``The impact of early dementia diagnosis and
  intervention on informal caregivers,'' \emph{Progress in neurobiology}, vol.
  110, pp. 54--62, 2013.

\bibitem{rasmussen2019alzheimer}
J.~Rasmussen and H.~Langerman, ``Alzheimer’s disease--why we need early
  diagnosis,'' \emph{Degenerative Neurological and Neuromuscular Disease},
  vol.~9, p. 123, 2019.

\bibitem{folstein1975mini}
M.~F. Folstein, S.~E. Folstein, and P.~R. McHugh, ``“mini-mental state”: a
  practical method for grading the cognitive state of patients for the
  clinician,'' \emph{Journal of psychiatric research}, vol.~12, no.~3, pp.
  189--198, 1975.

\bibitem{mohs1997development}
R.~C. Mohs, D.~Knopman, R.~C. Petersen, S.~H. Ferris, C.~Ernesto, M.~Grundman,
  M.~Sano, L.~Bieliauskas, D.~Geldmacher, C.~Clark \emph{et~al.}, ``Development
  of cognitive instruments for use in clinical trials of antidementia drugs:
  additions to the alzheimer's disease assessment scale that broaden its
  scope.'' \emph{Alzheimer disease and associated disorders}, 1997.

\bibitem{KAUFMAN2017105}
\BIBentryALTinterwordspacing
D.~M. Kaufman, H.~L. Geyer, and M.~J. Milstein, ``Chapter 7 - dementia,'' in
  \emph{Kaufman's Clinical Neurology for Psychiatrists (Eighth Edition)},
  eighth edition~ed., D.~M. Kaufman, H.~L. Geyer, and M.~J. Milstein,
  Eds.\hskip 1em plus 0.5em minus 0.4em\relax Elsevier, 2017, pp. 105--149.
  [Online]. Available:
  \url{https://www.sciencedirect.com/science/article/pii/B9780323415590000071}
\BIBentrySTDinterwordspacing

\bibitem{skinner2012alzheimer}
J.~Skinner, J.~O. Carvalho, G.~G. Potter, A.~Thames, E.~Zelinski, P.~K. Crane,
  L.~E. Gibbons, A.~D.~N. Initiative \emph{et~al.}, ``The alzheimer’s disease
  assessment scale-cognitive-plus (adas-cog-plus): an expansion of the adas-cog
  to improve responsiveness in mci,'' \emph{Brain imaging and behavior},
  vol.~6, no.~4, pp. 489--501, 2012.

\bibitem{ansart2020predicting}
M.~Ansart, S.~Epelbaum, G.~Bassignana, A.~B{\^o}ne, S.~Bottani, T.~Cattai,
  R.~Couronne, J.~Faouzi, I.~Koval, M.~Louis \emph{et~al.}, ``Predicting the
  progression of mild cognitive impairment using machine learning: a
  systematic, quantitative and critical review,'' \emph{Medical Image
  Analysis}, p. 101848, 2020.

\bibitem{utsumil2018personalized}
Y.~Utsumil, O.~O. Rudovicl, K.~Petersonl, R.~Guerrero, and R.~W. Picardl,
  ``Personalized gaussian processes for forecasting of alzheimer’s disease
  assessment scale-cognition sub-scale (adas-cog13),'' in \emph{2018 40th
  Annual International Conference of the IEEE Engineering in Medicine and
  Biology Society (EMBC)}.\hskip 1em plus 0.5em minus 0.4em\relax IEEE, 2018,
  pp. 4007--4011.

\bibitem{zhu2016canonical}
X.~Zhu, H.-I. Suk, S.-W. Lee, and D.~Shen, ``Canonical feature selection for
  joint regression and multi-class identification in alzheimer’s disease
  diagnosis,'' \emph{Brain imaging and behavior}, vol.~10, no.~3, pp. 818--828,
  2016.

\bibitem{prakash2020quantitative}
M.~Prakash, M.~Abdelaziz, L.~Zhang, B.~A. Strange, J.~Tohka, A.~D.~N.
  Initiative \emph{et~al.}, ``Quantitative longitudinal predictions of
  alzheimer’s disease by multi-modal predictive learning,'' \emph{Journal of
  Alzheimer's Disease}, no. Preprint, pp. 1--14, 2020.

\bibitem{tsao2017feature}
S.~Tsao, N.~Gajawelli, J.~Zhou, J.~Shi, J.~Ye, Y.~Wang, and N.~Lepor{\'e},
  ``Feature selective temporal prediction of alzheimer's disease progression
  using hippocampus surface morphometry,'' \emph{Brain and behavior}, vol.~7,
  no.~7, p. e00733, 2017.

\bibitem{wittenberg2019economic}
R.~Wittenberg, M.~Knapp, M.~Karagiannidou, J.~Dickson, and J.~M. Schott,
  ``Economic impacts of introducing diagnostics for mild cognitive impairment
  alzheimer's disease patients,'' \emph{Alzheimer's \& Dementia: Translational
  Research \& Clinical Interventions}, vol.~5, pp. 382--387, 2019.

\bibitem{vemuri2010role}
P.~Vemuri and C.~R. Jack, ``Role of structural mri in alzheimer's disease,''
  \emph{Alzheimer's research \& therapy}, vol.~2, no.~4, pp. 1--10, 2010.

\bibitem{wang2010high}
Y.~Wang, Y.~Fan, P.~Bhatt, and C.~Davatzikos, ``High-dimensional pattern
  regression using machine learning: from medical images to continuous clinical
  variables,'' \emph{Neuroimage}, vol.~50, no.~4, pp. 1519--1535, 2010.

\bibitem{jie2016temporally}
B.~Jie, M.~Liu, J.~Liu, D.~Zhang, and D.~Shen, ``Temporally constrained group
  sparse learning for longitudinal data analysis in alzheimer's disease,''
  \emph{IEEE Transactions on Biomedical Engineering}, vol.~64, no.~1, pp.
  238--249, 2016.

\bibitem{cao2018L2}
P.~Cao, X.~Liu, J.~Yang, D.~Zhao, M.~Huang, and O.~Zaiane, ``L2, 1- l1
  regularized nonlinear multi-task representation learning based cognitive
  performance prediction of alzheimer’s disease,'' \emph{Pattern
  Recognition}, vol.~79, pp. 195--215, 2018.

\bibitem{cao2019feature}
P.~Cao, S.~Tang, M.~Huang, J.~Yang, D.~Zhao, A.~Trabelsi, and O.~Zaiane,
  ``Feature-aware multi-task feature learning for predicting cognitive outcomes
  in alzheimer's disease,'' in \emph{2019 IEEE International Conference on
  Bioinformatics and Biomedicine (BIBM)}.\hskip 1em plus 0.5em minus
  0.4em\relax IEEE, 2019, pp. 1--5.

\bibitem{lei2020deep}
B.~Lei, M.~Yang, P.~Yang, F.~Zhou, W.~Hou, W.~Zou, X.~Li, T.~Wang, X.~Xiao, and
  S.~Wang, ``Deep and joint learning of longitudinal data for alzheimer's
  disease prediction,'' \emph{Pattern Recognition}, vol. 102, p. 107247, 2020.

\bibitem{bhagwat2019artificial}
N.~Bhagwat, J.~Pipitone, A.~N. Voineskos, M.~M. Chakravarty, A.~D.~N.
  Initiative \emph{et~al.}, ``An artificial neural network model for clinical
  score prediction in alzheimer disease using structural neuroimaging
  measures,'' \emph{Journal of psychiatry \& neuroscience: JPN}, vol.~44,
  no.~4, p. 246, 2019.

\bibitem{jiang2018correlation}
P.~Jiang, X.~Wang, Q.~Li, L.~Jin, and S.~Li, ``Correlation-aware sparse and
  low-rank constrained multi-task learning for longitudinal analysis of
  alzheimer's disease,'' \emph{IEEE journal of biomedical and health
  informatics}, vol.~23, no.~4, pp. 1450--1456, 2018.

\bibitem{huang2016longitudinal}
L.~Huang, Y.~Jin, Y.~Gao, K.-H. Thung, D.~Shen, A.~D.~N. Initiative
  \emph{et~al.}, ``Longitudinal clinical score prediction in alzheimer's
  disease with soft-split sparse regression based random forest,''
  \emph{Neurobiology of aging}, vol.~46, pp. 180--191, 2016.

\bibitem{zhou2013modeling}
J.~Zhou, J.~Liu, V.~A. Narayan, J.~Ye, A.~D.~N. Initiative \emph{et~al.},
  ``Modeling disease progression via multi-task learning,'' \emph{NeuroImage},
  vol.~78, pp. 233--248, 2013.

\bibitem{stonnington2010predicting}
C.~M. Stonnington, C.~Chu, S.~Kl{\"o}ppel, C.~R. Jack~Jr, J.~Ashburner, R.~S.
  Frackowiak, A.~D.~N. Initiative \emph{et~al.}, ``Predicting clinical scores
  from magnetic resonance scans in alzheimer's disease,'' \emph{Neuroimage},
  vol.~51, no.~4, pp. 1405--1413, 2010.

\bibitem{zhou2011multi}
J.~Zhou, L.~Yuan, J.~Liu, and J.~Ye, ``A multi-task learning formulation for
  predicting disease progression,'' in \emph{Proceedings of the 17th ACM SIGKDD
  international conference on Knowledge discovery and data mining}, 2011, pp.
  814--822.

\bibitem{yan2015cortical}
J.~Yan, T.~Li, H.~Wang, H.~Huang, J.~Wan, K.~Nho, S.~Kim, S.~L. Risacher, A.~J.
  Saykin, L.~Shen \emph{et~al.}, ``Cortical surface biomarkers for predicting
  cognitive outcomes using group l2, 1 norm,'' \emph{Neurobiology of aging},
  vol.~36, pp. S185--S193, 2015.

\bibitem{lu2020predicting}
L.~Lu, H.~Wang, S.~Elbeleidy, and F.~Nie, ``Predicting cognitive declines using
  longitudinally enriched representations for imaging biomarkers,'' in
  \emph{Proceedings of the IEEE/CVF Conference on Computer Vision and Pattern
  Recognition}, 2020, pp. 4827--4836.

\bibitem{duchesne2009relating}
S.~Duchesne, A.~Caroli, C.~Geroldi, D.~L. Collins, and G.~B. Frisoni,
  ``Relating one-year cognitive change in mild cognitive impairment to baseline
  mri features,'' \emph{Neuroimage}, vol.~47, no.~4, pp. 1363--1370, 2009.

\bibitem{jalali2013dirty}
A.~Jalali, P.~Ravikumar, and S.~Sanghavi, ``A dirty model for multiple sparse
  regression,'' \emph{IEEE Transactions on Information Theory}, vol.~59,
  no.~12, pp. 7947--7968, 2013.

\bibitem{tabarestani2020distributed}
S.~Tabarestani, M.~Aghili, M.~Eslami, M.~Cabrerizo, A.~Barreto, N.~Rishe, R.~E.
  Curiel, D.~Loewenstein, R.~Duara, and M.~Adjouadi, ``A distributed multitask
  multimodal approach for the prediction of alzheimer’s disease in a
  longitudinal study,'' \emph{NeuroImage}, vol. 206, p. 116317, 2020.

\bibitem{lei2017longitudinal}
B.~Lei, F.~Jiang, S.~Chen, D.~Ni, and T.~Wang, ``Longitudinal analysis for
  disease progression via simultaneous multi-relational temporal-fused
  learning,'' \emph{Frontiers in aging neuroscience}, vol.~9, p.~6, 2017.

\bibitem{wang2020modeling}
L.~Wang, L.~Xu, P.~Li, S.~Zha, and L.~Chen, ``Modeling disease progression via
  weakly supervised temporal multitask matrix completion,'' in \emph{2020 IEEE
  International Conference on Systems, Man, and Cybernetics (SMC)}.\hskip 1em
  plus 0.5em minus 0.4em\relax IEEE, 2020, pp. 1141--1148.

\bibitem{moradi2017predicting}
E.~Moradi, B.~Khundrakpam, J.~D. Lewis, A.~C. Evans, and J.~Tohka, ``Predicting
  symptom severity in autism spectrum disorder based on cortical thickness
  measures in agglomerative data,'' \emph{Neuroimage}, vol. 144, pp. 128--141,
  2017.

\bibitem{johnson2007adjusting}
W.~E. Johnson, C.~Li, and A.~Rabinovic, ``Adjusting batch effects in microarray
  expression data using empirical bayes methods,'' \emph{Biostatistics},
  vol.~8, no.~1, pp. 118--127, 2007.

\bibitem{manjon2010adaptive}
J.~V. Manj{\'o}n, P.~Coup{\'e}, L.~Mart{\'\i}-Bonmat{\'\i}, D.~L. Collins, and
  M.~Robles, ``Adaptive non-local means denoising of mr images with spatially
  varying noise levels,'' \emph{Journal of Magnetic Resonance Imaging},
  vol.~31, no.~1, pp. 192--203, 2010.

\bibitem{rajapakse1997statistical}
J.~C. Rajapakse, J.~N. Giedd, and J.~L. Rapoport, ``Statistical approach to
  segmentation of single-channel cerebral mr images,'' \emph{IEEE transactions
  on medical imaging}, vol.~16, no.~2, pp. 176--186, 1997.

\bibitem{tohka2004fast}
J.~Tohka, A.~Zijdenbos, and A.~Evans, ``Fast and robust parameter estimation
  for statistical partial volume models in brain mri,'' \emph{Neuroimage},
  vol.~23, no.~1, pp. 84--97, 2004.

\bibitem{ashburner2007fast}
J.~Ashburner, ``A fast diffeomorphic image registration algorithm,''
  \emph{Neuroimage}, vol.~38, no.~1, pp. 95--113, 2007.

\bibitem{rosen1984new}
W.~G. Rosen, R.~C. Mohs, and K.~L. Davis, ``A new rating scale for alzheimer's
  disease.'' \emph{The American journal of psychiatry}, 1984.

\bibitem{kueper2018alzheimer}
J.~K. Kueper, M.~Speechley, and M.~Montero-Odasso, ``The alzheimer’s disease
  assessment scale--cognitive subscale (adas-cog): modifications and
  responsiveness in pre-dementia populations. a narrative review,''
  \emph{Journal of Alzheimer's Disease}, vol.~63, no.~2, pp. 423--444, 2018.

\bibitem{fortin2017harmonization}
J.-P. Fortin, D.~Parker, B.~Tun{\c{c}}, T.~Watanabe, M.~A. Elliott, K.~Ruparel,
  D.~R. Roalf, T.~D. Satterthwaite, R.~C. Gur, R.~E. Gur \emph{et~al.},
  ``Harmonization of multi-site diffusion tensor imaging data,''
  \emph{Neuroimage}, vol. 161, pp. 149--170, 2017.

\bibitem{argyriou2008convex}
A.~Argyriou, T.~Evgeniou, and M.~Pontil, ``Convex multi-task feature
  learning,'' \emph{Machine learning}, vol.~73, no.~3, pp. 243--272, 2008.

\bibitem{ji2009accelerated}
S.~Ji and J.~Ye, ``An accelerated gradient method for trace norm
  minimization,'' in \emph{Proceedings of the 26th annual international
  conference on machine learning}, 2009, pp. 457--464.

\bibitem{obozinski2010joint}
G.~Obozinski, B.~Taskar, and M.~I. Jordan, ``Joint covariate selection and
  joint subspace selection for multiple classification problems,''
  \emph{Statistics and Computing}, vol.~20, no.~2, pp. 231--252, 2010.

\bibitem{zhou2011malsar}
J.~Zhou, J.~Chen, and J.~Ye, ``Malsar: Multi-task learning via structural
  regularization,'' \emph{Arizona State University}, vol.~21, 2011.

\bibitem{tibshirani1996regression}
R.~Tibshirani, ``Regression shrinkage and selection via the lasso,''
  \emph{Journal of the Royal Statistical Society: Series B (Methodological)},
  vol.~58, no.~1, pp. 267--288, 1996.

\bibitem{argyriou2007multi}
A.~Argyriou, T.~Evgeniou, and M.~Pontil, ``Multi-task feature learning,'' in
  \emph{Advances in neural information processing systems}, 2007, pp. 41--48.

\bibitem{liu2012multi}
J.~Liu, S.~Ji, and J.~Ye, ``Multi-task feature learning via efficient l2,
  1-norm minimization,'' \emph{arXiv preprint arXiv:1205.2631}, 2012.

\bibitem{nie2010efficient}
F.~Nie, H.~Huang, X.~Cai, and C.~H. Ding, ``Efficient and robust feature
  selection via joint l2, 1-norms minimization,'' in \emph{Advances in neural
  information processing systems}, 2010, pp. 1813--1821.

\bibitem{jalali2010dirty}
A.~Jalali, S.~Sanghavi, C.~Ruan, and P.~K. Ravikumar, ``A dirty model for
  multi-task learning,'' in \emph{Advances in neural information processing
  systems}, 2010, pp. 964--972.

\bibitem{he2016novel}
D.~He, D.~Kuhn, and L.~Parida, ``Novel applications of multitask learning and
  multiple output regression to multiple genetic trait prediction,''
  \emph{Bioinformatics}, vol.~32, no.~12, pp. i37--i43, 2016.

\bibitem{fazel2003matrix}
S.~M. Fazel, ``Matrix rank minimization with applications.'' 2003.

\bibitem{fortin2018harmonization}
J.-P. Fortin, N.~Cullen, Y.~I. Sheline, W.~D. Taylor, I.~Aselcioglu, P.~A.
  Cook, P.~Adams, C.~Cooper, M.~Fava, P.~J. McGrath \emph{et~al.},
  ``Harmonization of cortical thickness measurements across scanners and
  sites,'' \emph{Neuroimage}, vol. 167, pp. 104--120, 2018.

\bibitem{orlhac2018postreconstruction}
F.~Orlhac, S.~Boughdad, C.~Philippe, H.~Stalla-Bourdillon, C.~Nioche,
  L.~Champion, M.~Soussan, F.~Frouin, V.~Frouin, and I.~Buvat, ``A
  postreconstruction harmonization method for multicenter radiomic studies in
  pet,'' \emph{Journal of Nuclear Medicine}, vol.~59, no.~8, pp. 1321--1328,
  2018.

\bibitem{yu2018statistical}
M.~Yu, K.~A. Linn, P.~A. Cook, M.~L. Phillips, M.~McInnis, M.~Fava, M.~H.
  Trivedi, M.~M. Weissman, R.~T. Shinohara, and Y.~I. Sheline, ``Statistical
  harmonization corrects site effects in functional connectivity measurements
  from multi-site fmri data,'' \emph{Human brain mapping}, vol.~39, no.~11, pp.
  4213--4227, 2018.

\bibitem{good2006permutation}
P.~I. Good, \emph{Permutation, parametric, and bootstrap tests of
  hypotheses}.\hskip 1em plus 0.5em minus 0.4em\relax Springer Science \&
  Business Media, 2006.

\bibitem{lewis2018t1}
J.~D. Lewis, A.~C. Evans, J.~Tohka, B.~D.~C. Group \emph{et~al.}, ``T1
  white/gray contrast as a predictor of chronological age, and an index of
  cognitive performance,'' \emph{Neuroimage}, vol. 173, pp. 341--350, 2018.

\bibitem{huttunen2012meg}
H.~Huttunen, T.~Manninen, and J.~Tohka, ``Meg mind reading: strategies for
  feature selection,'' \emph{Proc. Fed. Comput. Sci. Event}, vol. 2012, pp.
  42--49, 2012.

\bibitem{qian2013glmnet}
J.~Qian, T.~Hastie, J.~Friedman, R.~Tibshirani, and N.~Simon, ``Glmnet for
  matlab,'' \emph{Accessed: Nov}, vol.~13, no. 2017, pp. 4--2, 2013.

\bibitem{chang2011libsvm}
C.-C. Chang and C.-J. Lin, ``Libsvm: A library for support vector machines,''
  \emph{ACM transactions on intelligent systems and technology (TIST)}, vol.~2,
  no.~3, pp. 1--27, 2011.

\bibitem{moradi2015machine}
E.~Moradi, A.~Pepe, C.~Gaser, H.~Huttunen, J.~Tohka, A.~D.~N. Initiative
  \emph{et~al.}, ``Machine learning framework for early mri-based alzheimer's
  conversion prediction in mci subjects,'' \emph{Neuroimage}, vol. 104, pp.
  398--412, 2015.

\bibitem{wang2019multi}
M.~Wang, D.~Zhang, D.~Shen, and M.~Liu, ``Multi-task exclusive relationship
  learning for alzheimer’s disease progression prediction with longitudinal
  data,'' \emph{Medical image analysis}, vol.~53, pp. 111--122, 2019.

\bibitem{zandifar2020mri}
A.~Zandifar, V.~S. Fonov, S.~Ducharme, S.~Belleville, D.~L. Collins, A.~D.~N.
  Initiative \emph{et~al.}, ``Mri and cognitive scores complement each other to
  accurately predict alzheimer's dementia 2 to 7 years before clinical onset,''
  \emph{NeuroImage: Clinical}, vol.~25, p. 102121, 2020.

\end{thebibliography}

 \begin{IEEEbiography}[{\includegraphics[width=1in,height=1.25in,clip,keepaspectratio]{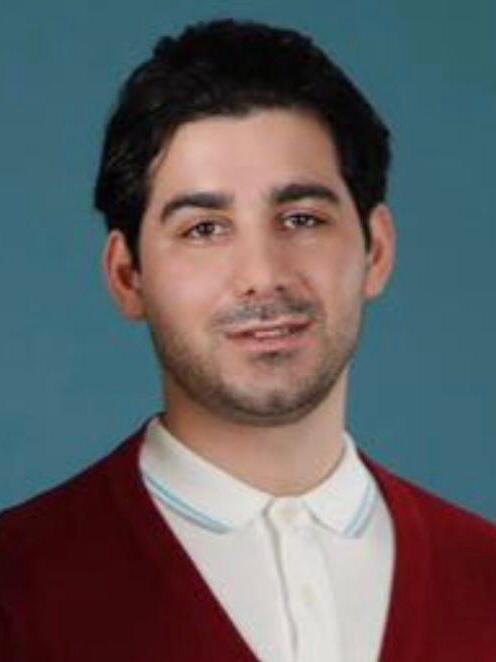}}]{Vandad Imani} received a master’s degree in artificial intelligence from the Central University of Hyderabad, India, in 2014. He is currently pursuing the Ph.D. degree with the A. I. Virtanen Institute for Molecular Sciences, University of Eastern Finland, Kuopio, Finland. His research interests include machine learning, pattern recognition, and their applications to predict brain disorders.
 \vspace{0mm}
\end{IEEEbiography}

 \begin{IEEEbiography}[{\includegraphics[width=1in,height=1.25in,clip,keepaspectratio]{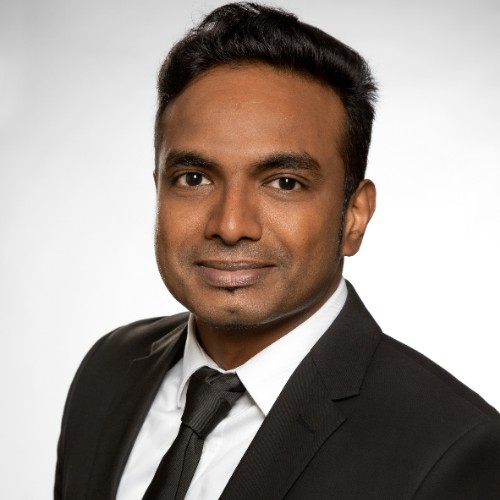}}]{MITHILESH PRAKASH} was born in Bellary, India, in 1988. He received a B.E in Medical Electronics from V.T.U, India in 2009, an M.Sc. (Tech) in Biomedical Engineering from Tampere University of Technology, Finland in 2015, and a Ph.D. degree in Applied Physics from the University of Eastern Finland, Finland in 2019. He has over 4 years of Software programming industry experience. He is currently working as a Postdoctoral Researcher at the A. I. Virtanen Institute for Molecular Sciences, University of Eastern Finland, Kuopio, Finland. His research interests include machine learning, algorithm designing, and programming for their applications in Biomedical Engineering. 

\end{IEEEbiography}

 \begin{IEEEbiography}[{\includegraphics[width=1in,height=1.25in,clip,keepaspectratio]{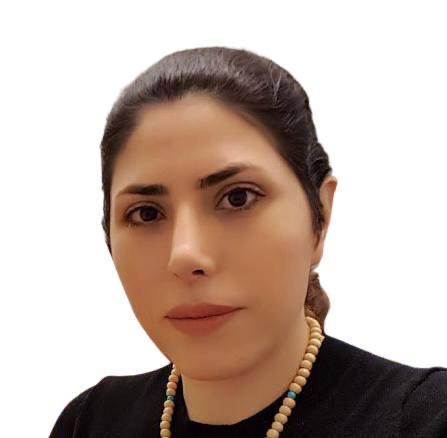}}]{MARZIEH ZARE} received her M.S. degrees in the Signal Processing department from Tampere University, Tampere, Finland, in 2017. She was also a research assistant at the A. I. Virtanen Institute for Molecular Sciences, University of Eastern Finland, Kuopio, Finland with supervision of Prof. Jussi Tohka, in 2017-2018.
She is currently a Ph.D. degree student in Computing Sciences at Tampere University, Tampere, Finland. Her research interests include data analysis, Signal processing, machine learning, and pattern recognition and their applications to industrial solutions. 
\end{IEEEbiography}

\begin{IEEEbiography}[{\includegraphics[width=1in,height=1.25in,clip,keepaspectratio]{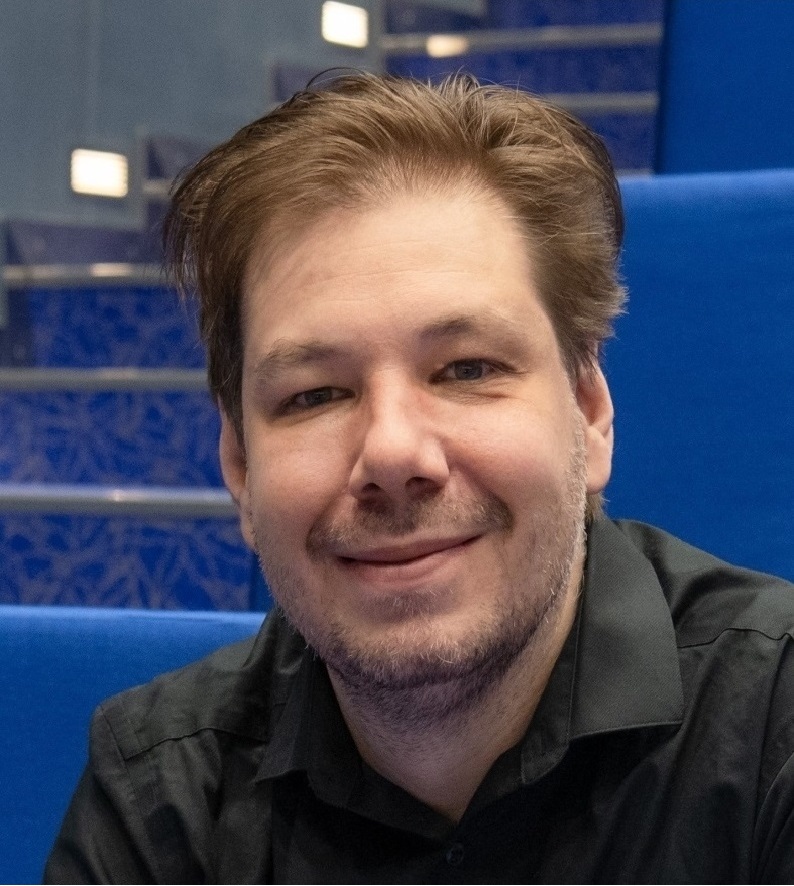}}]{Jussi Tohka} received a Ph.D. in signal processing from the Tampere University of Technology, Finland, in 2003. He was a Postdoctoral Fellow with the University of California at Los Angeles, Los Angeles, USA. He held an Academy Research Fellow position at the Department of Signal Processing, Tampere University of Technology. He was a CONEX Professor with the Department of Bioengineering and Aerospace Engineering, Universidad Carlos III de Madrid, Spain. He is currently with the A. I. Virtanen Institute for Molecular Sciences, University of Eastern Finland, Kuopio, Finland. His research interests include machine learning, image analysis, and pattern recognition and their applications to brain imaging.

\end{IEEEbiography}
\EOD
\end{document}